\documentclass[12pt, titlepage, reqno]{article}
\usepackage{authblk}
\usepackage[margin=1in]{geometry}
\RequirePackage{amsthm,amsmath,amsfonts,amssymb}
\usepackage{array}
\RequirePackage{natbib}
\RequirePackage[colorlinks,citecolor=blue,urlcolor=blue]{hyperref}
\usepackage{booktabs,longtable}
\usepackage{tabularx}
\usepackage{multirow}
\usepackage{threeparttable}
\usepackage{graphicx}
\graphicspath{{./}{../figures/}}
\usepackage{xcolor}
\usepackage{soul}
\usepackage{epstopdf}
\usepackage[inline]{enumitem}
\usepackage{bm}
\usepackage[ruled, lined]{algorithm2e}
\usepackage{setspace}
\usepackage{tikz}
\usetikzlibrary{arrows.meta,positioning,calc,decorations.pathreplacing}

\newcommand\pkg[1]{\texttt{#1}}
\let\proglang=\textsf 

\title{Statistical, Machine Learning and Deep Learning Methods for 
Missing Covariates: An Empirical Evaluation with Applications to 
Alzheimer's Disease Research}

\author[1]{Yiqing Pan}

\author[2]{Sharon X.\ Xie}

\author[3]{Shujie Ma}

\author[4,5,6,7]{Angela L.\ Jefferson}

\author[1,4,5,*]{Panpan Zhang}

\affil[1]{Department of Biostatistics, Vanderbilt University Medical 
Center, Nashville, TN, USA}

\affil[2]{Department of Biostatistics, Epidemiology, and Informatics, 
University of Pennsylvania, Philadelphia, PA, USA}

\affil[3]{Department of Statistics,	University of California, 
Riverside, CA, USA}

\affil[4]{Vanderbilt Memory \& Alzheimer's Center, Vanderbilt 
University Medical Center, Nashville, TN, USA}

\affil[5]{Vanderbilt Alzheimer's Disease Research Center, Vanderbilt 
University Medical Center, Nashville, TN, USA}

\affil[6]{Department of Neurology, Vanderbilt University Medical 
Center, Nashville, TN, USA}

\affil[7]{Department of Medicine, Vanderbilt University Medical 
Center, Nashville, TN, USA}

\affil[*]{Correspondence: 
\href{mailto:panpan.zhang@vumc.org}{Panpan Zhang}}

\begin{document}
	
\maketitle
	
\begin{abstract}

Missing data are common in biomedical research, particularly when 
biomarkers are costly or require invasive procedures to obtain, such 
as cerebrospinal fluid measures in Alzheimer's disease. Although a 
wide range of missing data methods exists, from traditional 
statistical approaches to modern machine learning (ML) and deep 
learning (DL) algorithms, their comparative performance under 
different missingness mechanisms remains insufficiently understood. 
In this study, we provide an empirical evaluation of eight methods 
for handling missing covariates, including complete-case analysis, 
multiple imputation, ML-based techniques, and DL frameworks such as 
multilayer perceptrons, variational autoencoders, and generative 
adversarial networks. We summarize their conceptual foundations, 
practical implementation considerations, strengths, and limitations. 
Through extensive simulations, we assess estimation bias, efficiency, 
standard error accuracy, confidence interval coverage, and 
computational burden across different missingness mechanisms and 
proportions. An application to the Vanderbilt Memory and Aging 
Project further illustrates the implications of different methods for 
regression analysis in a real-world biomedical setting. Finally, we 
provide practical recommendations for method selection and highlight 
remaining challenges that warrant future investigation.


\end{abstract}

\doublespacing

\section{Introduction}
\label{sec:intro}

Missing data are common in biomedical research, arising from factors 
such as the high cost of biomarker evaluation and the invasiveness of 
certain clinical procedures. A notable example in Alzheimer's disease 
(AD) research is cerebrospinal fluid (CSF) biomarkers. Although they 
are critical indicators of AD pathology, CSF data often undergo high 
rates of missingness because data acquisition through lumbar puncture 
is invasive and may impose substantial burdens on participants. Such 
missingness may lead to the exclusion of participants with incomplete 
data from statistical analyses, thereby reducing statistical power 
and potentially compromising the validity of inference. If not 
properly addressed, missing data can undermine the reliability of 
research findings~\citep{allison2002missing, little2019statistical}.

Missing data are classified into three major categories 
\citep{rubin1976inference}: missing completely at random (MCAR), 
missing at random (MAR), and missing not at random (MNAR), where 
MCAR and MAR are considered ignorable for likelihood-based 
inference \citep{little2019statistical}. Methods for missing data 
have improved vastly in the last several decades, from the simplest 
complete case analysis to advanced methods, such as multiple 
imputation, maximum likelihood estimation, inverse probability 
weighting, and weighted generalized estimating equations, among 
others~\citep{ibrahim2005missing, ibrahim2009missing}. Interested 
readers are referred to~\citet{pigott2001areview, schafer2002missing, 
ibrahim2009missing, sterne2009multiple} for classical and modern 
reviews on missing data methods in various disciplines.

With the rapid development of machine learning (ML) in the past two 
decades, ML has become particularly prominent in the field of data 
science, including solving the missing data 
problems~\citep{emmanuel2021asurvey}. ML methods offer advantages in 
capturing complex, nonlinear relationships and interactions among 
variables without imposing restrictive functional 
forms~\citep{liu2022greedy, givens2023density, lim2024deeply}. In 
addition, ML models are often more scalable than traditional methods 
for large datasets.  However, their predictive flexibility does not 
necessarily ensure valid inference under MNAR mechanisms, which 
generally require additional assumptions or explicit modeling of the 
missingness process. Deep learning (DL), as one of the most critical 
subsets of ML, has only recently begun to be used for missing data 
problems. Compared with classic ML models, DL algorithms exploit more 
advanced methods to characterize complex relationships between 
variables and are more applicable to high-dimensional and massive 
data sets. In particular, generative models in DL, such as generative 
adversarial networks, create new samples by modeling the underlying 
data distribution, thereby providing a novel imputation strategy.

Despite substantial methodological advances, the comparative 
performance of traditional statistical, ML, and DL methods for 
missing covariate imputation remains insufficiently understood, 
particularly with respect to downstream statistical inference. To 
address this gap, we conduct an empirical evaluation of eight 
representative methods under different missingness mechanisms and 
proportions, focusing on estimation bias, efficiency, and inferential 
reliability. Rather than providing an exhaustive review of all 
available approaches, our study aims to establish a common evaluation 
framework and offer practical guidance for method selection in 
biomedical research. This work is motivated by the Vanderbilt 
Memory and Aging Project~\citep[VMAP,][]{osborn2018cerebrospinal}, in 
which more than 50\% of CSF biomarker data are missing at baseline 
due to the invasiveness and participant burden of lumbar puncture. 
This kind of missing covariates is particularly consequential, as 
they usually reduce statistical power, bias associations between 
biomarkers and cognition, and ultimately limit the translational 
impact of AD research. We therefore use VMAP both to guide the design 
of our simulation study and as a case study to demonstrate how 
different imputation strategies perform in a real-world biomedical 
setting. We want to highlight that, unlike a purely 
narrative review, our work combines extensive simulations with a 
comprehensive examination of available software implementations in 
\proglang{R} programming language, and we provide example codes in an 
open resource to facilitate direct application by other researchers.

The remainder of this manuscript is organized as follows.
Section~\ref{sec:method} details the six ML and DL imputation 
methods, including methodological concepts and implementation 
details. Section~\ref{sec:sim} presents a comprehensive simulation 
study examining a range of missingness mechanisms and missing 
proportions. Section~\ref{sec:app} demonstrates the application of 
selected methods to the VMAP cohort and compares their performance in 
a real-world setting. Finally, Section~\ref{sec:dis} concludes with a 
summary of key findings, limitations, and directions for future 
research. 

\section{Methods}
\label{sec:method}

In this section, we review six prominent ML and DL methods that are 
adaptable for missing data imputation. Namely, they are $k$-nearest 
neighbors (kNN), support vector machine (SVM), Random Forest (RF), 
Multilayer Perceptron (MLP), generative adversarial networks (GAN), 
and Variational Autoencoder (VAE). For each surveyed method, we 
will provide a general overview of the algorithm, demonstrate how it 
can be applied to missing data imputation, and discuss its strengths 
and limitations. Lastly, we will also introduce the available 
\proglang{R} package that can be used for implementation.

These methods differ not only in their underlying algorithms but also 
in whether they produce single or multiple imputations. This 
distinction is particularly relevant to downstream statistical 
inference, as the treatment of imputation uncertainty may affect the 
estimation of standard errors and confidence interval coverage.

\subsection{$K$-Nearest Neighbors (KNN)}
\label{sec:knn}

\begin{figure}
	\centering
	\includegraphics[width=0.9\textwidth]{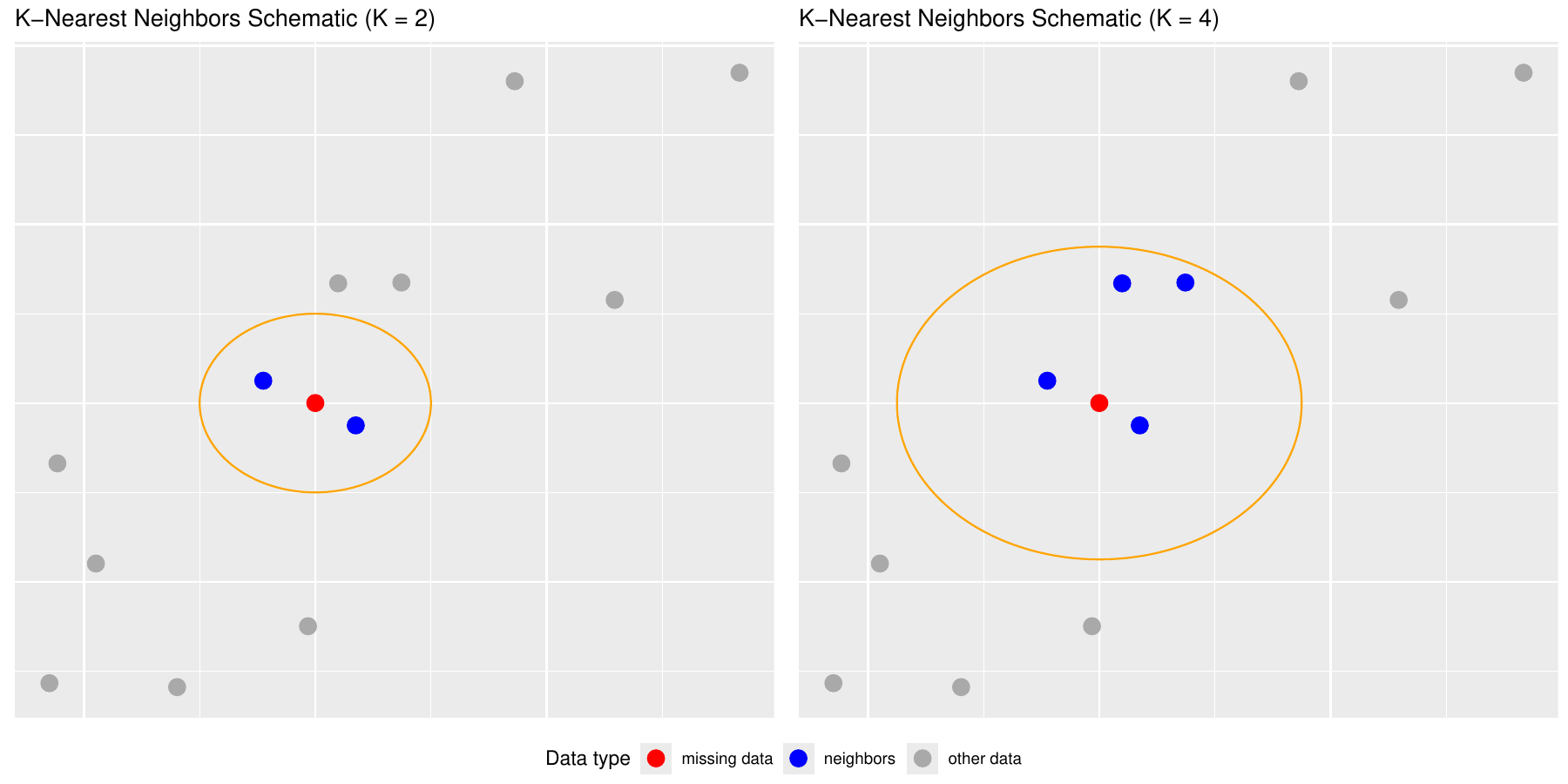}
	\caption{Examples of kNN schematic.}
	\label{fig:knn_schematic}
\end{figure}

The $k$-nearest neighbors (kNN) algorithm is a versatile and widely 
used method in machine learning for classification tasks 
\citep{fix1989discriminatory, Garcia2009knn}. It assigns a label to a 
data point based on the class information among its $k$ nearest 
neighbors, where ``closeness'' is typically measured using a 
distance metric, such as Euclidean distance (for continuous 
variables) and matching (for categorical variables). An important 
extension of kNN is its use in missing data 
imputation~\citep{zhang2012nearest, lalande2022numerical}. In this 
context, the classical kNN algorithm identifies the nearest neighbors 
of missing 
data and imputes the missing entries by averaging (for continuous 
variables) or selecting the most frequent value (for categorical 
variables) from these neighbors. This approach leverages the inherent 
relationships in the data, making it a promising tool for dealing 
with missing values while preserving the data structure. Schematic 
examples of kNN are given in Figure~\ref{fig:knn_schematic}.

The effectiveness of kNN for imputation is dependent on the choice of 
three key factors: the distance metric, the neighborhood size $k$, 
and the adaptation method, each of which will be discussed 
subsequently. In kNN, the most commonly used distance metrics for 
continuous variables are Euclidean and Manhattan distances, both of 
which are special cases of Minkowski distance, while for 
categorical variables, matching and non-matching are typically 
considered. When dealing with a mix of continuous and categorical 
variables, advanced distance metrics like Gower's 
distance~\citep{gower1971ageneral} and grey 
distance~\citep{zhang2012nearest, huang2016grey} are often used. In 
practice, the \pkg{VIM} package~\citep{pkg:VIM} utilizes Gower's 
distance, whereas other packages, such as 
\pkg{DMwR2}~\citep{pkg:DMwR2} and \pkg{kknn}~\citep{pkg:kknn}, rely 
on traditional distance metrics.

The selection of $k$ varies across studies, with some choosing a 
specified value based on expert knowledge and others utilizing all 
available neighbors. In the absence of prior information, 
\citet{lall1996anearest} proposed setting $k = \sqrt{n_0}$ for $n_0 > 
100$, where $n_0$ is the number of observations in the dataset. 
A more advanced algorithm for optimally selecting $k$ was introduced 
by~\citet{samworth2012optimal}, which has been implemented in 
the~\pkg{kknn} package. In contrast, the \pkg{VIM} and \pkg{DMwR2} 
packages do not include specific methods for selecting $k$, but a 
cross-validation (CV) approach may be employed instead. CV is 
directly available in the~\pkg{kknn} package, and can be combined 
the build-in functions from the \pkg{VIM} and \pkg{DMwR2} 
packages by manual encoding.

The weighted average is among the most commonly used adaption methods 
for kNN. In this approach, missing values are imputed using the 
weighted average of their selected $k$ nearest neighbors, where the 
weights are typically inversely proportional to the distances. This 
method is implemented in the \pkg{VIM} package. In contrast, the 
\pkg{DMwR2} package performs $k$-nearest neighbors imputation without 
incorporating weights. The \pkg{kknn} package employs a kernel-based 
method to determine weights based on the theoretical framework 
developed by \citet{samworth2012optimal}. A summary of \proglang{R} 
packages for imputation using kNN is given in Table~\ref{tab:knnR}.

\begin{table}[tbp]
	\centering
	\small 
	\setlength{\tabcolsep}{12.3pt}
	\begin{tabular}{l r r r}
		\toprule
		& \pkg{VIM} & \pkg{DMwR2} & \pkg{kknn} \\
		\midrule
		Imputation & Built-in function & Built-in function & 
		Manual encoding \\
		Data type & Continuous, categorical & Continuous, categorical 
		& Continuous \\
		\multirow{2}{*}{Distance} & Continuous: Euclidean & 
		Continuous: Euclidean & \multirow{2}{*}{Minkowski} \\
		& Categorical: Gower & Categorical: Hamming & \\
		$K$ selection & Pre-specified & Pre-specified & 
		Kernel, CV \\
		Adaption & Weighted (distance) & Unweighted & Weighted 
		(kernel) \\
		\bottomrule		
	\end{tabular}
		\caption{Summary of \proglang{R} packages for imputation 
		using kNN.}
	\label{tab:knnR}
\end{table}

Although kNN is a classical algorithm, it remains widely used for 
missing data imputation due to its conceptual simplicity and 
nonparametric nature. The method primarily relies on neighboring 
observations, rendering it to effectively capture local structure in 
the data. However, kNN can be computationally intensive for large 
datasets, as it requires calculating distances between each instance 
with missing values and all other data points to identify the $k$ 
nearest neighbors~\citep{batista2003KNN}. The method is also 
sensitive to the choice of $k$ and distance metric, which may affect 
the stability and reproducibility of the results. Moreover, kNN 
typically uses all available covariates, including irrelevant or 
noisy ones, potentially degrading imputation performance. Finally, 
because kNN imputes one variable at a time, it may struggle to 
capture dependencies among variables when there exist multiple 
missing variables.

\subsection{Support Vector Machine (SVM)}

\begin{figure}[tbp]
	\centering
	\includegraphics[width=0.9\textwidth]{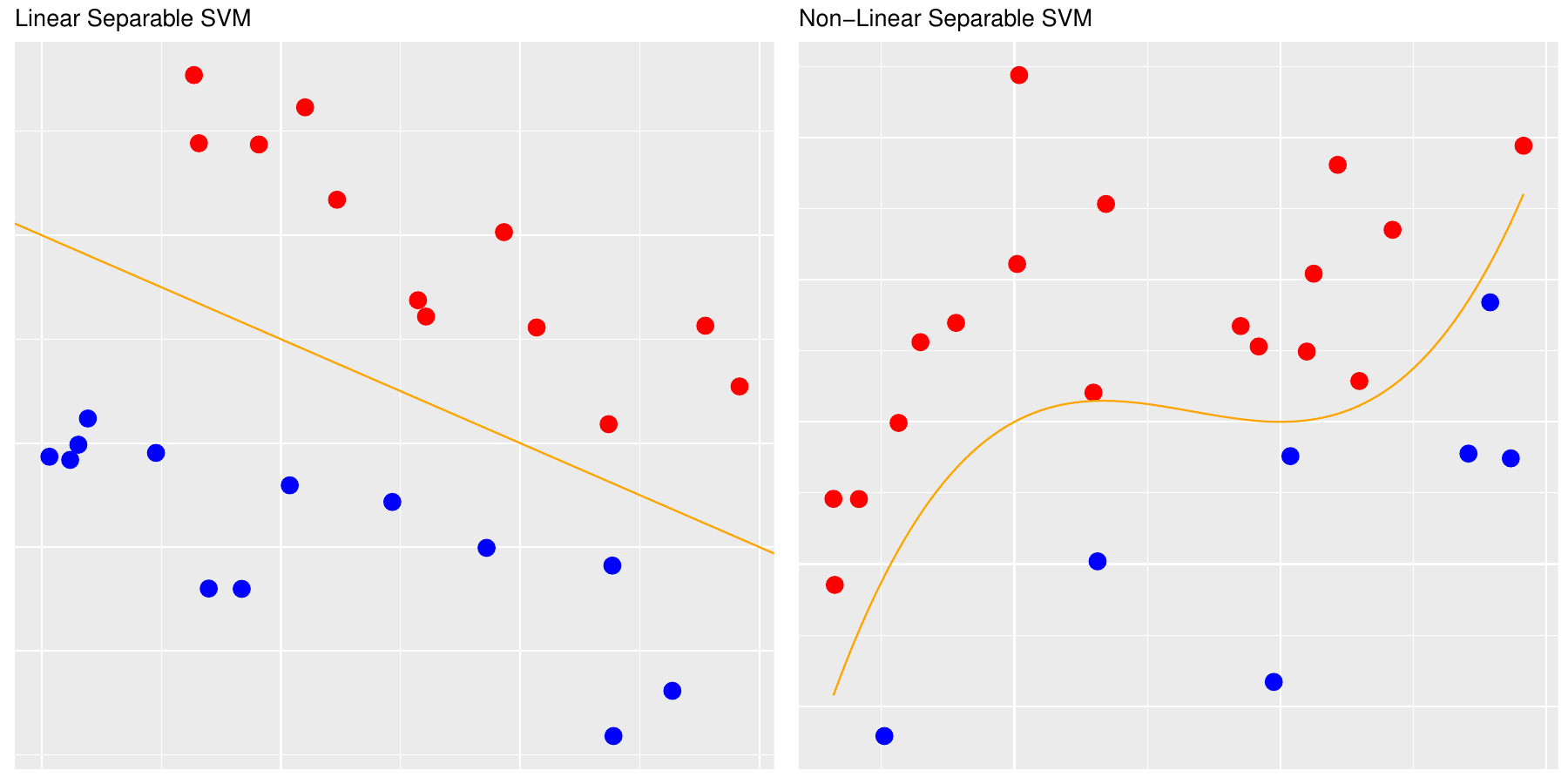}
	\caption{Examples of linearly separable and nonlinearly separable 
		SVM schematics.}
	\label{fig:svm_schematic}
\end{figure}

Support Vector Machine (SVM) is a powerful supervised learning 
algorithm that identifies an optimal hyperplane to effectively 
separate data points into distinct classes~\citep{cortes1995support}. 
The primary goal of SVM is to determine a decision boundary that 
maximizes the margin between the classes, where the margin is 
defined as the distance between the hyperplane and the nearest data 
points (i.e., support vectors). As a result, SVM is formulated as 
an optimization problem. For linearly separable data (e.g., the left 
panel of Figure~\ref{fig:svm_schematic}), the hard-margin SVM 
enforces strict classification constraints using Lagrange 
multipliers~\citep{kecman2005support}. In contrast, for nonlinearly 
separable data (e.g., the right panel of 
Figure~\ref{fig:svm_schematic}), the soft-margin SVM 
introduces a regularization parameter to balance margin maximization 
and classification error minimization. It is worth noting that SVM 
was originally developed as a binary classifier. Extensions such as 
one-vs-one and one-vs-rest strategies have been proposed to 
accommodate multi-class classification 
tasks~\citep{galar2011anoverview}.

\begin{figure}[tbp]
	\centering
	\includegraphics[width=0.6\textwidth]{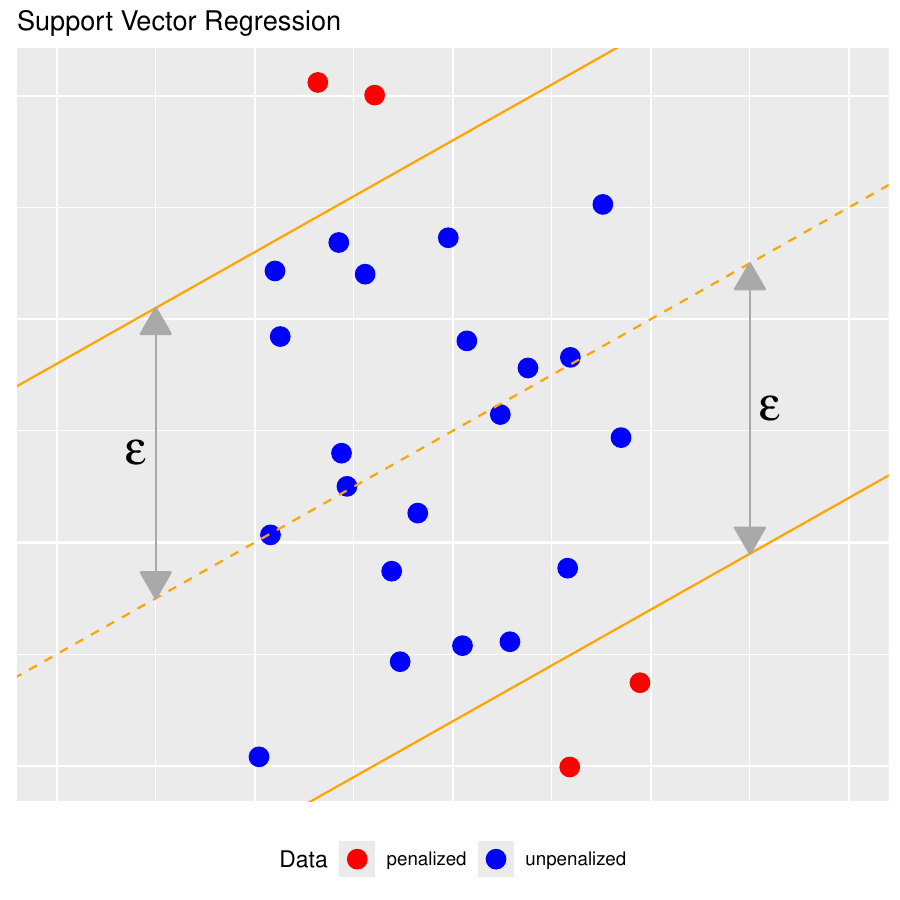}
	\caption{An example of $\varepsilon$-insensitive tube for SVR.}
	\label{fig:svr_schematic}
\end{figure}

Support Vector Regression (SVR) is an extension of SVM designed for 
regression tasks. It aims to find a linear function that closely 
approximates target values while prioritizing model simplicity and 
robustness to outliers. SVR introduces an $\varepsilon$-insensitive 
loss function, where deviations within the threshold (i.e., 
$\varepsilon$) are ignored, visualized as an 
$\varepsilon$-insensitive tube around the regression function; see 
Figure~\ref{fig:svr_schematic} as an example.
Predictions falling outside this tube are penalized, allowing the 
model to tolerate small errors and reduce sensitivity to 
noise. A larger $\varepsilon$ increases the robustness of SVR to 
outliers by ignoring larger deviations. 

When regularization is applied in SVR, the regularization parameter 
$C$ which penalizes deviations of predictions from the 
$\varepsilon$-insensitive tube, is 
used to govern the trade-off between model complexity and tolerance 
to prediction errors. In the literature, $C$ is also known as the 
cost parameter. A greater value of $C$ imposes a higher penalty 
on errors, thus prioritizing a model fitting the training data more 
closely. To handle nonlinear relationships, SVR leverages the kernel 
trick, implicitly mapping inputs into a higher-dimensional space 
where a linear function can be fitted. Among the most commonly used 
kernels is the Radial Basis Function (RBF) kernel, whose shape is 
governed by a tuning parameter $\sigma$. A smaller value of $\sigma$ 
results in a narrower kernel so that the model becomes more sensitive 
to individual training points. While the kernel trick enables SVR to 
handle complex, nonlinearly separable data efficiently, a linear 
kernel may be preferable when the underlying relationship is 
approximately linear to improve interpretability, and meanwhile, 
reduce computational cost.

SVM and SVR have been utilized for missing data imputation due to 
their ability to model complex, nonlinear relationships between 
variables~\citep{stewart2018constructing}. In this context, SVM is 
typically employed for imputing categorical variables, while 
SVR is used for continuous variables. The imputation process involves 
training the model on observed data to learn the underlying 
functional relationship between available features and the variable 
with missing values. Then, the trained SVM or SVR models can be 
exploited to predict missing data leveraging the learned structure.

In practice, SVM and SVR can be implemented in \proglang{R} using the 
\pkg{e1071}~\citep{pkg:e1071}, \pkg{kernlab}~\citep{pkg:kernlab}, and 
\pkg{caret}~\citep{pkg:caret} packages. While none of these packages 
offers built-in imputation functions, user-defined implementations 
are straightforward and compatible with both categorical and 
continuous variables. All three packages support standard kernels 
(such as linear, polynomial, RBF) for modeling nonlinearity. However, 
only \pkg{caret} supports automated tuning of the cost parameter $C$ 
via CV. In contrast, \pkg{e1071} and \pkg{kernlab} 
require pre-specified parameters with fixed defaults ($\varepsilon = 
0.1$, $C = 1$) and do not include tuning routines. None of these 
packages directly supports tuning the parameter $\varepsilon$. 
Table~\ref{tab:SVMR} provides a concise comparison of the three 
packages. Through our investigation, incorporating \pkg{kernlab} 
within the \pkg{SuperLearner} framework~\citep{pkg:SuperLearner}, a 
comprehensive ensemble learning library, appears to be a promising 
approach for optimally selecting tuning parameters. However, the 
methodological details fall beyond the scope of the present paper, 
and are therefore omitted.

\begin{table}[tbp]
	\centering
	\small 
	\setlength{\tabcolsep}{12.7pt}
	\begin{tabular}{l r r r}
		\toprule
		& \pkg{e1071} & \pkg{kernlab} & \pkg{caret} \\
		\midrule
		Imputation & Manual encoding & Manual encoding & Manual 
		encoding \\
		Support & SVM, SVR & SVM, SVR & SVM, SVR \\
		Tuning Parameter & Pre-specified & Pre-specified & Automation 
		\\
		Kernel & Available & Available & Available \\
		\bottomrule		
	\end{tabular}
		\caption{Summary of \proglang{R} packages for imputation 
		using SVM.}
	\label{tab:SVMR}
\end{table}

SVM and SVR offer several advantages for missing data imputation, 
especially in scenarios involving complex, nonlinear relationships or 
high-dimensional data. By utilizing kernel functions, SVM and SVR can 
capture intricate data structures without specifying explicit 
transformations, which makes them well-suited for imputing values in 
datasets with complex relationships~\citep{smola2004atutorial}. 
Additionally, the margin-based optimization nature makes SVM and SVR 
robust to overfitting, even in high-dimensional spaces. However, the 
performance of these methods is highly sensitive to the choices of 
tuning parameters and kernel function. Moreover, SVM and SVR can be 
computationally intensive for large datasets, as they involve solving 
a quadratic optimization problem. Another limitation is that SVM is 
often considered a ``black box,'' with limited transparency regarding 
the internal mechanisms behind imputations, making the interpretation 
of results challenging.

\subsection{Random Forest (RF)}
\label{sec:rf}

Classification and regression trees (CARTs) impute missing values by 
recursively partitioning the dataset into subsets based on 
explanatory variables. At each internal node, the data is split 
according to a decision rule on one of the predictors, and the 
process continues until reaching terminal nodes (i.e., leaves) that 
contain the final, homogeneous subsets of observations. This produces 
a single decision tree that models the relationship between 
predictors and the variable with missing data, allowing missing 
values to be imputed from patterns observed in the complete cases. 
CART is capable of handling both categorical and continuous 
variables, making it a versatile imputation tool for datasets with 
mixed data types and complex missingness patterns. Random forest (RF) 
extends this approach by replacing a single tree with an ensemble of 
trees, each built on a bootstrap sample of the data and a randomly 
selected subset of predictors at each 
split~\citep{breiman2001random}. This 
design 
reduces overfitting, stabilizes predictions, and better reflects 
uncertainty by capturing multiple plausible relationships between 
variables. In the multiple imputation context, RF inherently produces 
variation across imputations, while its non-parametric nature allows 
it to model nonlinearities and complex interactions without strong 
distributional assumptions~\citep{stekhoven2011missingforest}. As 
discussed by~\citet{tang2017random}, RF-based imputation can be 
implemented using three main strategies, each leveraging these 
ensemble properties to improve accuracy and robustness.

\begin{figure}[tbp]
	\centering
	\begin{tikzpicture}
		\node[circle, draw = black, minimum size = 0.5cm] (C11) at 
		(0, 0) {}; 
		\node[circle, draw = black, minimum size = 0.5cm] (C12) at 
		(-1, -1) {};
		\node[circle, draw = black, minimum size = 0.5cm] (C13) at 
		(1, 
		-1) {};
		\node[fill = blue, minimum size = 0.5cm] (T11) at (-1.5, -2) 
		{};
		\node[fill = blue, minimum size = 0.5cm] (T12) at (-0.5, -2) 
		{};
		\node[fill = blue, minimum size = 0.5cm] (T13) at (0.5, -2) 
		{};
		\node[fill = blue, minimum size = 0.5cm] (T14) at (1.5, -2) 
		{};
		\draw (C11) -- (C12); 
		\draw (C11) -- (C13);
		\draw (C12) -- (T11);
		\draw (C12) -- (T12);
		\draw (C13) -- (T13);
		\draw (C13) -- (T14);
		\node[circle, draw = black, minimum size = 0.5cm] (C21) at 
		(5, 0) {}; 
		\node[circle, draw = black, minimum size = 0.5cm] (C22) at 
		(4, -1) {};
		\node[fill = blue, minimum size = 0.5cm] (T21) at (6, -1) {};
		\node[fill = blue, minimum size = 0.5cm] (T22) at (3.5, -2) 
		{};
		\node[circle, draw = black, minimum size = 0.5cm] (C23) at 
		(4.5, -2) {};
		\node[fill = blue, minimum size = 0.5cm] (T23) at (4, -3) {};
		\node[fill = blue, minimum size = 0.5cm] (T24) at (5, -3) {};
		\draw (C21) -- (C22); 
		\draw (C21) -- (T21);
		\draw (C22) -- (T22);
		\draw (C22) -- (C23);
		\draw (C23) -- (T23);
		\draw (C23) -- (T24);
		\node[circle, draw = black, minimum size = 0.5cm] (C31) at 
		(10, 0) {}; 
		\node[circle, draw = black, minimum size = 0.5cm] (C32) at 
		(9, -1) {};
		\node[circle, draw = black, minimum size = 0.5cm] (C33) at 
		(11, -1) {};
		\node[fill = blue, minimum size = 0.5cm] (T31) at (8.5, -2) 
		{};
		\node[fill = blue, minimum size = 0.5cm] (T32) at (9.5, -2) 
		{};
		\node[circle, draw = black, minimum size = 0.5cm] (C34) at 
		(11.5, -2) {};
		\node[fill = blue, minimum size = 0.5cm] (T33) at (11.5, -2) 
		{};
		\node[fill = blue, minimum size = 0.5cm] (T34) at (11, -3) {};
		\node[fill = blue, minimum size = 0.5cm] (T35) at (12, -3) {};
		\draw (C31) -- (C32); 
		\draw (C31) -- (C33);
		\draw (C32) -- (T31);
		\draw (C32) -- (T32);
		\draw (C33) -- (C34);
		\draw (C33) -- (T33);
		\draw (C34) -- (T34);
		\draw (C34) -- (T35);
	\end{tikzpicture}
	\caption{Example of a random forest consisting of three decision 
		trees; white circles and blue squares represent internal 
		nodes 
		and terminal nodes, respectively.}
	\label{fig:rf}
\end{figure}
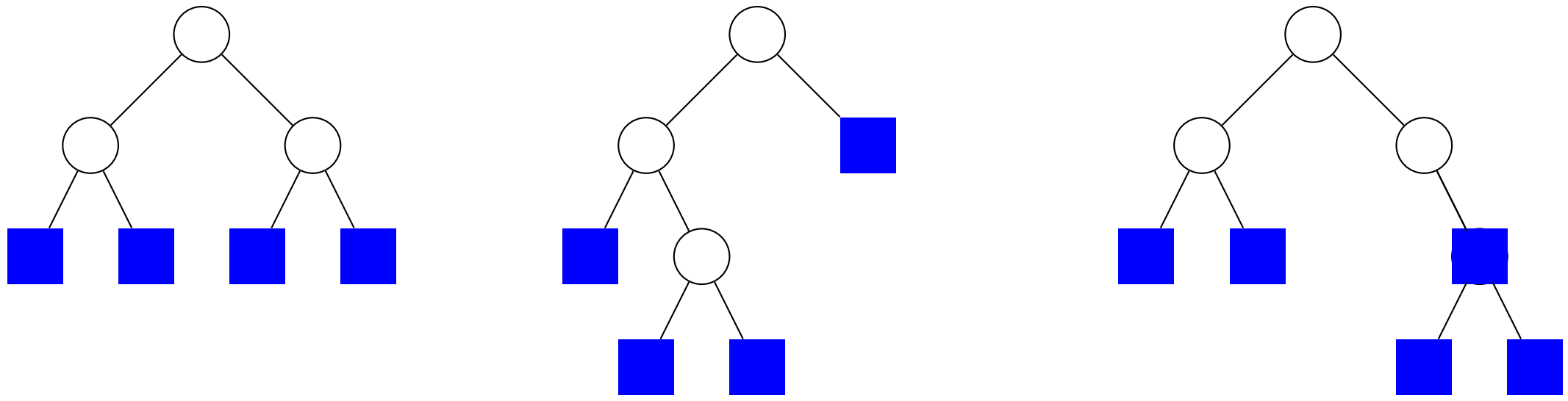

The first strategy, originally proposed by \citet{breiman2001random} 
begins 
with an initial imputation of the missing values using some ad-hoc 
approaches (e.g., mean substitution), followed by fitting a random 
forest to the complete data. A proximity matrix is then computed 
from the forest, measuring the frequency of pairs of observations 
falling into the same terminal node across trees. Missing values are 
updated using proximity-weighted information from their most similar 
observed cases. This process is repeated iteratively until the 
imputed values converge.

The second strategy integrates imputation directly into the 
forest-growing process, without a distinct pre-imputation phase. 
Missing values are handled on-the-fly during tree construction, for 
example by sending cases with missing split variables down multiple 
branches with fractional weights or by using surrogate splits. 
Predictions for missing entries are generated as part of the forest 
building, and the process is iterated to refine imputations until 
stability is reached.

The third strategy starts with an initial imputation, then 
sequentially focuses on each variable with missing data. For each 
such variable, a random forest is trained using all other variables 
as predictors and only rows where the target variable is observed. 
The trained model is used to predict and update the missing values 
for that variable. This per-variable modeling cycle continues across 
all incomplete variables, and the entire sequence is iterated until 
the changes between successive imputations are negligible.

\begin{table}[tbp]
  \centering
  \small
  \setlength{\tabcolsep}{10.5pt}
  \begin{tabular}{l r r r r}
  	\toprule
	& \pkg{missForest} & \pkg{missRanger} & \pkg{mice::rf} & 
	\pkg{miceRanger} \\
	\midrule
	Imputation Type & Single & Single & Multiple & Multiple \\
	Core RF Engine & \pkg{randomForest} & \pkg{ranger} & 
	\pkg{randomForest} & \pkg{ranger} \\ 
	Scalability & Low & High & Low & High \\
	Mixed-Type Data & Yes & Yes & Yes & Yes \\
	Diagnostics & OOB error & Limited & MICE diagnostics & Limited \\
	\bottomrule
  \end{tabular}
  \caption{Summary of \proglang{R} packages for imputation 
		using RF.}
  \label{tab:rfimpute}
\end{table}
    
Most existing RF-based imputation tools follow a nonparametric, 
iterative strategy for imputing mixed-type data. 
Two typical \proglang{R} packages are
\pkg{missForest}~\citep{pkg:missForest} and 
\pkg{missRanger}~\citep{pkg:missRanger}. While 
\pkg{missForest} is built on the 
\pkg{randomForest} engine~\citep{pkg:randomForest}, and is known for 
robust accuracy and 
out-of-bag~\citep[OOB,][]{janitza2018on} error estimates, it suffers 
from slow computation 
and high memory demand on large datasets. In contrast, 
\pkg{missRanger} adopts 
the more efficient \pkg{ranger} engine~\citep{pkg:ranger}, providing 
substantial gains 
in speed and scalability. However, both packages are limited to 
single imputation, reducing their appeal for modern inference 
analyses.

The other two prevalent packages, 
\pkg{mice::rf}~\citep{vanbuuren2011mice} and 
\pkg{miceRanger}~\citep{pkg:miceranger}, support 
multiple imputation workflows. Specifically, \pkg{mice::rf} embeds RF 
imputation within the multiple imputation by chained equations (MICE) 
framework but inherits slower computation from the legacy 
\pkg{randomForest} backend. In comparison, \pkg{miceRanger} 
implements chained RFs using \pkg{ranger}, achieving greater 
computational efficiency and improved handling of large, 
high-dimensional datasets. Table~\ref{tab:rfimpute} summarizes and 
compares the main features of these RF-based packages.

\subsection{Multilayer Perceptrons}
\label{sec:mlp}

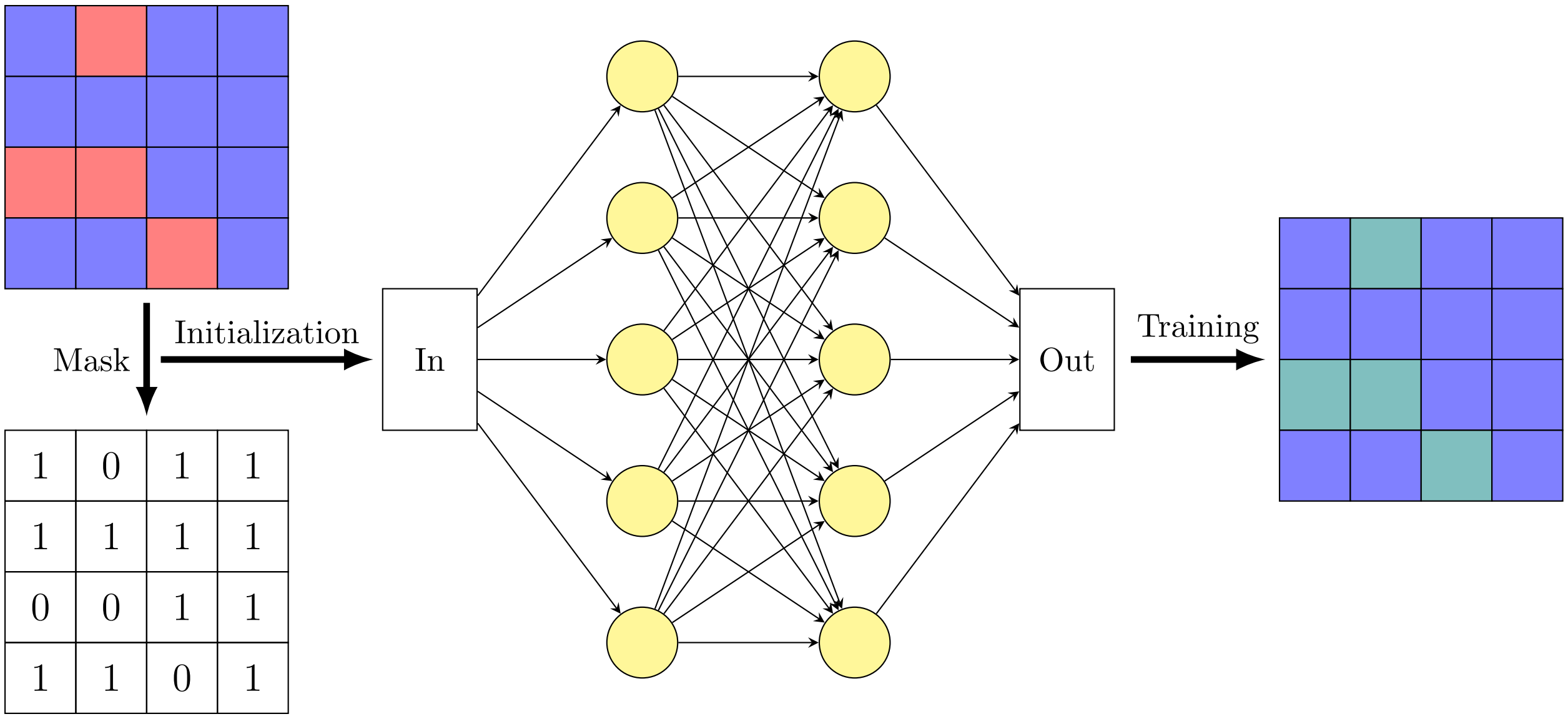
\begin{figure}
	\centering
	\begin{tikzpicture}[scale=0.75]
		\draw[draw = black, fill = blue!50] (0, 6) rectangle (1, 5);
		\draw[draw = black, fill = red!50] (1, 6) rectangle (2, 5);
		\draw[draw = black, fill = blue!50] (2, 6) rectangle (3, 5);
		\draw[draw = black, fill = blue!50] (3, 6) rectangle (4, 5);
		\draw[draw = black, fill = blue!50] (0, 5) rectangle (1, 4);
		\draw[draw = black, fill = blue!50] (1, 5) rectangle (2, 4);
		\draw[draw = black, fill = blue!50] (2, 5) rectangle (3, 4);
		\draw[draw = black, fill = blue!50] (3, 5) rectangle (4, 4);
		\draw[draw = black, fill = red!50] (0, 4) rectangle (1, 3);
		\draw[draw = black, fill = red!50] (1, 4) rectangle (2, 3);
		\draw[draw = black, fill = blue!50] (2, 4) rectangle (3, 3);
		\draw[draw = black, fill = blue!50] (3, 4) rectangle (4, 3);
		\draw[draw = black, fill = blue!50] (0, 3) rectangle (1, 2);
		\draw[draw = black, fill = blue!50] (1, 3) rectangle (2, 2);
		\draw[draw = black, fill = red!50] (2, 3) rectangle (3, 2);
		\draw[draw = black, fill = blue!50] (3, 3) rectangle (4, 2);
		
		\draw (0, 0) rectangle node{$1$} (1, -1);
		\draw (1, 0) rectangle node{$0$} (2, -1);
		\draw (2, 0) rectangle node{$1$} (3, -1);
		\draw (3, 0) rectangle node{$1$} (4, -1);
		\draw (0, -1) rectangle node{$1$} (1, -2);
		\draw (1, -1) rectangle node{$1$} (2, -2);
		\draw (2, -1) rectangle node{$1$} (3, -2);
		\draw (3, -1) rectangle node{$1$} (4, -2);
		\draw (0, -2) rectangle node{$0$} (1, -3);
		\draw (1, -2) rectangle node{$0$} (2, -3);
		\draw (2, -2) rectangle node{$1$} (3, -3);
		\draw (3, -2) rectangle node{$1$} (4, -3);
		\draw (0, -3) rectangle node{$1$} (1, -4);
		\draw (1, -3) rectangle node{$1$} (2, -4);
		\draw (2, -3) rectangle node{$0$} (3, -4);
		\draw (3, -3) rectangle node{$1$} (4, -4);
		
		\draw[-latex, line width = 2pt] (2, 1.8)--(2, 0.2)
		node[midway, left, font = \footnotesize] {Mask};
		
		\node[rectangle, draw = black, minimum width = 1cm, minimum 
		height = 1.5cm, font = \footnotesize] (input) at (6, 1) 
		{In}; 
		
		\draw[-latex, line width = 2pt] (2.2, 1)--(5.2, 1)
		node[midway, above, font = \footnotesize] {Initialization};
				
		\node[circle, draw = black, fill = yellow!50, minimum size = 
		0.75cm] (L11) at (9, 5) {}; 
		\node[circle, draw = black, fill = yellow!50, minimum size = 
		0.75cm] (L12) at (9, 3) {}; 
		\node[circle, draw = black, fill = yellow!50, minimum size = 
		0.75cm] (L13) at (9, 1) {}; 
		\node[circle, draw = black, fill = yellow!50, minimum size = 
		0.75cm] (L14) at (9, -1) {};
		\node[circle, draw = black, fill = yellow!50, minimum size = 
		0.75cm] (L15) at (9, -3) {};  
		
		\draw[-stealth] (input) -- (L11);
		\draw[-stealth] (input) -- (L12);
		\draw[-stealth] (input) -- (L13);
		\draw[-stealth] (input) -- (L14);
		\draw[-stealth] (input) -- (L15);
		
		\node[circle, draw = black, fill = yellow!50, minimum size = 
		0.75cm] (L21) at (12, 5) {}; 
		\node[circle, draw = black, fill = yellow!50, minimum size = 
		0.75cm] (L22) at (12, 3) {}; 
		\node[circle, draw = black, fill = yellow!50, minimum size = 
		0.75cm] (L23) at (12, 1) {}; 
		\node[circle, draw = black, fill = yellow!50, minimum size = 
		0.75cm] (L24) at (12, -1) {};
		\node[circle, draw = black, fill = yellow!50, minimum size = 
		0.75cm] (L25) at (12, -3) {}; 
		
		\draw[-stealth] (L11) -- (L21);
		\draw[-stealth] (L11) -- (L22);
		\draw[-stealth] (L11) -- (L23);
		\draw[-stealth] (L11) -- (L24);
		\draw[-stealth] (L11) -- (L25);
		\draw[-stealth] (L12) -- (L21);
		\draw[-stealth] (L12) -- (L22);
		\draw[-stealth] (L12) -- (L23);
		\draw[-stealth] (L12) -- (L24);
		\draw[-stealth] (L12) -- (L25);
		\draw[-stealth] (L13) -- (L21);
		\draw[-stealth] (L13) -- (L22);
		\draw[-stealth] (L13) -- (L23);
		\draw[-stealth] (L13) -- (L24);
		\draw[-stealth] (L13) -- (L25);
		\draw[-stealth] (L14) -- (L21);
		\draw[-stealth] (L14) -- (L22);
		\draw[-stealth] (L14) -- (L23);
		\draw[-stealth] (L14) -- (L24);
		\draw[-stealth] (L14) -- (L25);
		\draw[-stealth] (L15) -- (L21);
		\draw[-stealth] (L15) -- (L22);
		\draw[-stealth] (L15) -- (L23);
		\draw[-stealth] (L15) -- (L24);
		\draw[-stealth] (L15) -- (L25);
		
		\node[rectangle, draw = black, minimum width = 1cm, minimum 
		height = 1.5cm, font = \footnotesize] (output) at (15, 1) 
		{Out}; 
		
		\draw[-stealth] (L21) -- (output);
		\draw[-stealth] (L22) -- (output);
		\draw[-stealth] (L23) -- (output);
		\draw[-stealth] (L24) -- (output);
		\draw[-stealth] (L25) -- (output);
		
		\draw[-latex, line width = 2pt] (15.9, 1)--(17.8, 1)
		node[midway, above, font = \footnotesize] {Training};
		
		\draw[draw = black, fill = blue!50] (18, 3) rectangle (19, 2);
		\draw[draw = black, fill = teal!50] (19, 3) rectangle (20, 
		2);
		\draw[draw = black, fill = blue!50] (20, 3) rectangle (21, 2);
		\draw[draw = black, fill = blue!50] (21, 3) rectangle (22, 2);
		\draw[draw = black, fill = blue!50] (18, 2) rectangle (19, 1);
		\draw[draw = black, fill = blue!50] (19, 2) rectangle (20, 1);
		\draw[draw = black, fill = blue!50] (20, 2) rectangle (21, 1);
		\draw[draw = black, fill = blue!50] (21, 2) rectangle (22, 1);
		\draw[draw = black, fill = teal!50] (18, 1) rectangle (19, 
		0);
		\draw[draw = black, fill = teal!50] (19, 1) rectangle (20, 
		0);
		\draw[draw = black, fill = blue!50] (20, 1) rectangle (21, 0);
		\draw[draw = black, fill = blue!50] (21, 1) rectangle (22, 0);
		\draw[draw = black, fill = blue!50] (18, 0) rectangle (19, 
		-1);
		\draw[draw = black, fill = blue!50] (19, 0) rectangle (20, 
		-1);
		\draw[draw = black, fill = teal!50] (20, 0) rectangle (21, 
		-1);
		\draw[draw = black, fill = blue!50] (21, 0) rectangle (22, 
		-1);
	\end{tikzpicture}
	\caption{An example of MLP workflow: Left panel presents a 
	dataset (observed data in blue and missing data in red) and its 
	associated missingness indicator; Middle panel is an MLP diagram 
	(neurons in the hidden layers in yellow); Right panel represents 
	the imputed dataset (imputed data in green).}
	\label{fig:mlp}
\end{figure}

Multilayer Perceptrons (MLPs), one of the foundational and 
versatile architectures within deep learning, provide a flexible 
framework for modeling complex and nonlinear dependencies among 
variables without relying on any specific distributional 
assumptions~\citep{hornik1989multilayer, murtagh1991multilayer}. An 
MLP is structured as a feedforward artificial neural 
network, consisting of an input layer, several hidden layers 
composed of densely connected neurons, and an output layer. 
Typically, when employing an MLP for imputation tasks, the input 
layer is initialized with neurons equivalent to the number of 
features, while the output layer mimics this configuration to 
predict the complete feature set. However, variations exist by 
including additional neurons that may augment the input layer when 
auxiliary features are available to improve imputation accuracy. The 
hidden layers, positioned between the input layer and the output 
layer, are tasked with capturing intricate relationships within the 
dataset and may contain a large number of neurons depending on 
training needs. See Figure~\ref{fig:mlp} for a graphical illustration.

Each neuron in the hidden and output layers computes a weighted sum 
of its inputs, which is then transformed by a nonlinear activation 
function~\citep{kruse2022multilayer}. The rectified linear 
unit~\citep[ReLU,][]{nair2010rectified} is most widely adopted for 
its computational efficiency and favorable gradient properties, 
though alternatives such as the sigmoid or hyperbolic tangent 
functions remain useful in specific 
contexts~\citep{popescu2009multilayer, rezaeianzadeh2010prediction}. 
Model training is performed using 
backpropagation~\citep{rumelhart1986learning}, with 
weights and biases updated through stochastic gradient 
descent~\citep{amari1993backpropagation} or adaptive optimizers such 
as Adam~\citep{kingma2015adam}. 

When applied to missing data imputation, MLPs are trained to recover 
missing entries from patterns learned in the observed data. This task 
is typically formulated as supervised learning, with the model fit on 
either complete cases or artificially masked subsets of the observed 
data that serve as the training 
targets~\citep{silvaramirez2011missing, silvaramirez2015single}. The 
choice of loss function depends on the type of variable being 
imputed; for example, mean squared error (MSE) is commonly used for 
continuous outcomes, while cross-entropy is employed for categorical 
variables. For datasets containing mixed data types, customized or 
composite loss functions can be tailored to accommodate the 
characteristics of each variable. In practice, the training objective 
is commonly evaluated using artificially masked entries for which the 
true values are known. A binary mask identifies these training 
targets, allowing the loss function to measure reconstruction 
accuracy while excluding genuinely missing entries whose values are 
unavailable. Architecturally, simple feed-forward MLPs are prevalent, 
often structured with fully connected layers that enable them to 
model complex relationships between features, although recent work 
has also explored hybrid models that combine MLPs with autoencoder 
frameworks or regularization informed by domain 
knowledge~\citep{talwar2018autoimpute, qi2025mbsparse}. Model 
performance evaluation typically reports reconstruction error on 
held-out artificially masked entries, using metrics such as root mean 
squared error (RMSE) for continuous variables and classification 
accuracy for categorical outcomes, as well as overall downstream task 
performance based on the application scenario. Robust evaluation 
often requires multiple training-validation-testing splits or 
cross-validation to avoid overfitting and to provide stable estimates 
of imputation accuracy.

Implementing MLPs for missing data imputation in \proglang{R} relies 
on Python interface packages, most notably 
\pkg{keras3}~\citep{pkg:keras3} (or its legacy counterpart 
\pkg{keras}~\citep{pkg:keras}) in conjunction with 
\pkg{tensorflow}~\citep{pkg:tensorflow}, which provide high-level and 
low-level utilities, respectively, for building and training neural 
networks. The process typically begins with an initial ad hoc 
imputation (e.g., mean substitution) to provide a complete data 
matrix, along with the creation of indicator masks denoting missing 
entries. The imputation task is then formulated as regression for 
continuous variables or classification for categorical variables, 
with the MLPs trained to predict missing values based on observed 
features. Architectural design choices include the number of hidden 
layers, the number of neurons per layer, activation functions, and 
optimization algorithms. Once trained, the model can be applied 
iteratively to the dataset for prediction and subsequent training, 
and the process is repeated until imputations converge.

Theoretically, MLPs offer several compelling advantages for missing 
data imputation. Their layered neural network architecture and 
flexible activation functions help capture nonlinear relationships 
and complex interactions that traditional imputation models may fail, 
particularly in the presence of non-additive effects 
among predictors. They can be applied across diverse variable types 
and extended to multimodal or longitudinal datasets, making them 
especially appealing in biomedical research. However, research 
specifically evaluating MLPs for imputation remains 
limited~\citep{silvaramirez2011missing}, and these strengths come 
with important trade-offs. Effective training typically requires 
substantial amounts of complete data, with performance degrading in 
small-sample settings. The ``black-box'' nature of MLPs, combined 
with sensitivity to architectural and tuning choices (e.g., number of 
layers and units, learning rate schedules, and activation functions), 
highlights the need for careful model specification and rigorous 
validation. These challenges are amplified under MNAR mechanisms, 
where the missingness process itself must be accurately characterized 
to avoid biased imputations. Moreover, interpretability is limited 
compared with regression-based approaches, often necessitating 
complementary sensitivity analyses to support transparent reporting. 
Finally, training deep networks can be computationally intensive, 
which may constrain routine use in large-scale biomedical studies 
without adequate computing resources.

\subsection{Generative Adversarial Network (GAN)}

Generative adversarial 
networks~\citep[GANs,][]{goodfellow2014generative, 
goodfellow2020generative} are a class of deep generative models that 
approximate complex data distributions by training two neural 
networks in opposition within a minimax framework defined by a 
cross-entropy-based objective~\citep{creswell2018generative, 
pan2020loss}. Specifically, the generator, typically implemented as 
an MLP or similar architecture, transforms random noise or partially 
observed data into synthetic samples intended to resemble the real 
data. The discriminator, also a neural network, evaluates 
each sample and assesses whether it is genuine or generated. Through 
iterative adversarial training, the generator progressively improves 
its ability to produce realistic imputations, while the discriminator 
simultaneously enhances its capacity to distinguish true observations 
from synthetic values.

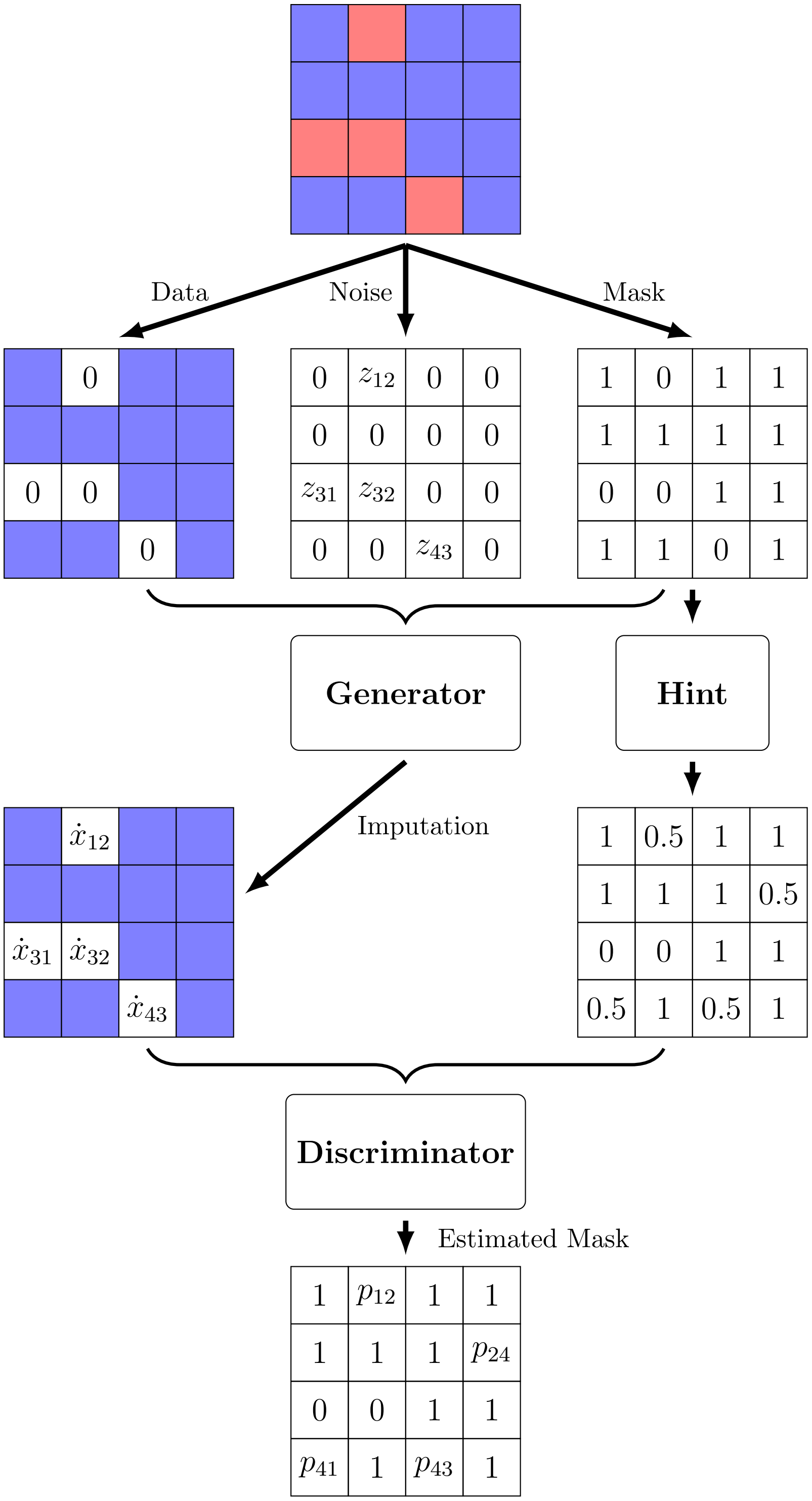
\begin{figure}
	\centering
	\begin{tikzpicture}[scale=0.75]
		\draw[draw = black, fill = blue!50] (0, 6) rectangle (1, 5);
		\draw[draw = black, fill = red!50] (1, 6) rectangle (2, 5);
		\draw[draw = black, fill = blue!50] (2, 6) rectangle (3, 5);
		\draw[draw = black, fill = blue!50] (3, 6) rectangle (4, 5);
		\draw[draw = black, fill = blue!50] (0, 5) rectangle (1, 4);
		\draw[draw = black, fill = blue!50] (1, 5) rectangle (2, 4);
		\draw[draw = black, fill = blue!50] (2, 5) rectangle (3, 4);
		\draw[draw = black, fill = blue!50] (3, 5) rectangle (4, 4);
		\draw[draw = black, fill = red!50] (0, 4) rectangle (1, 3);
		\draw[draw = black, fill = red!50] (1, 4) rectangle (2, 3);
		\draw[draw = black, fill = blue!50] (2, 4) rectangle (3, 3);
		\draw[draw = black, fill = blue!50] (3, 4) rectangle (4, 3);
		\draw[draw = black, fill = blue!50] (0, 3) rectangle (1, 2);
		\draw[draw = black, fill = blue!50] (1, 3) rectangle (2, 2);
		\draw[draw = black, fill = red!50] (2, 3) rectangle (3, 2);
		\draw[draw = black, fill = blue!50] (3, 3) rectangle (4, 2);
		
		\draw (0, 0) rectangle node{$0$} (1, -1);
		\draw (1, 0) rectangle node{$z_{12}$} (2, -1);
		\draw (2, 0) rectangle node{$0$} (3, -1);
		\draw (3, 0) rectangle node{$0$} (4, -1);
		\draw (0, -1) rectangle node{$0$} (1, -2);
		\draw (1, -1) rectangle node{$0$} (2, -2);
		\draw (2, -1) rectangle node{$0$} (3, -2);
		\draw (3, -1) rectangle node{$0$} (4, -2);
		\draw (0, -2) rectangle node{$z_{31}$} (1, -3);
		\draw (1, -2) rectangle	node{$z_{32}$} (2, -3);
		\draw (2, -2) rectangle node{$0$} (3, -3);
		\draw (3, -2) rectangle node{$0$} (4, -3);
		\draw (0, -3) rectangle node{$0$} (1, -4);
		\draw (1, -3) rectangle node{$0$} (2, -4);
		\draw (2, -3) rectangle node{$z_{43}$} (3, -4);
		\draw (3, -3) rectangle node{$0$} (4, -4);
		
		\draw (5, 0) rectangle node{$1$} (6, -1);
		\draw (6, 0) rectangle node{$0$} (7, -1);
		\draw (7, 0) rectangle node{$1$} (8, -1);
		\draw (8, 0) rectangle node{$1$} (9, -1);
		\draw (5, -1) rectangle node{$1$} (6, -2);
		\draw (6, -1) rectangle node{$1$} (7, -2);
		\draw (7, -1) rectangle node{$1$} (8, -2);
		\draw (8, -1) rectangle node{$1$} (9, -2);
		\draw (5, -2) rectangle node{$0$} (6, -3);
		\draw (6, -2) rectangle node{$0$} (7, -3);
		\draw (7, -2) rectangle node{$1$} (8, -3);
		\draw (8, -2) rectangle node{$1$} (9, -3);
		\draw (5, -3) rectangle node{$1$} (6, -4);
		\draw (6, -3) rectangle node{$1$} (7, -4);
		\draw (7, -3) rectangle node{$0$} (8, -4);
		\draw (8, -3) rectangle node{$1$} (9, -4);
		
		\draw[draw = black, fill = blue!50] (-5, 0) rectangle 
		node{} (-4, -1);
		\draw (-4, 0) rectangle node{$0$} (-3, -1);
		\draw[draw = black, fill = blue!50] (-3, 0) rectangle 
		node{} (-2, -1);
		\draw[draw = black, fill = blue!50] (-2, 0) rectangle 
		node{} (-1, -1);
		\draw[draw = black, fill = blue!50] (-5, -1) rectangle 
		node{} (-4, -2);
		\draw[draw = black, fill = blue!50] (-4, -1) rectangle 
		node{} (-3, -2);
		\draw[draw = black, fill = blue!50] (-3, -1) rectangle 
		node{} (-2, -2);
		\draw[draw = black, fill = blue!50] (-2, -1) rectangle 
		node{} (-1, -2);
		\draw (-5, -2) rectangle node{$0$} (-4, -3);
		\draw (-4, -2) rectangle node{$0$} (-3, -3);
		\draw[draw = black, fill = blue!50] (-3, -2) rectangle 
		node{} (-2, -3);
		\draw[draw = black, fill = blue!50] (-2, -2) rectangle 
		node{} (-1, -3);
		\draw[draw = black, fill = blue!50] (-5, -3) rectangle 
		node{} (-4, -4);
		\draw[draw = black, fill = blue!50] (-4, -3) rectangle 
		node{} (-3, -4);
		\draw (-3, -3) rectangle node{$0$} (-2, -4);
		\draw[draw = black, fill = blue!50] (-2, -3) rectangle 
		node{} (-1, -4);
		
		\draw[-latex, line width = 2pt] (2, 1.8)--(2, 0.2)
		node[midway, left, font = \footnotesize] {Noise};
		
		\draw[-latex, line width = 2pt] (2, 1.8)--(-3, 0.2)
		node[midway, left, xshift = -15pt, font = \footnotesize] 
		{Data};
		
		\draw[-latex, line width = 2pt] (2, 1.8)--(7, 0.2)
		node[midway, right, xshift = 15pt, font = \footnotesize] 
		{Mask};
		
		\draw[decorate, very thick, 
		decoration={brace, amplitude = 12pt, mirror}] 
		(-2.5, -4.2) -- (6.5, -4.2) node[midway] {};
		
		\node[draw, rounded corners = 3pt, minimum width = 3cm, 
		minimum height = 1.5cm] at (2, -6) {\textbf{Generator}};
		
		\draw[-latex, line width = 2pt] (7, -4.2)--(7, -4.8);
		
		\node[draw, rounded corners = 3pt, minimum width = 2cm, 
		minimum height = 1.5cm] at (7, -6) {\textbf{Hint}};
		
		\draw[-latex, line width = 2pt] (2, -7.2)--(-0.8, -9.5) 
		node[midway, right, xshift = 7pt, font = \footnotesize] 
		{Imputation};
		
		\draw[-latex, line width = 2pt] (7, -7.2)--(7, -7.8);
		
		\draw (-5, -8)[draw = black, fill = blue!50] rectangle 
		node{} (-4, -9);
		\draw (-4, -8) rectangle node{$\dot{x}_{12}$} (-3, -9);
		\draw (-3, -8)[draw = black, fill = blue!50] rectangle 
		node{} (-2, -9);
		\draw (-2, -8)[draw = black, fill = blue!50] rectangle 
		node{} (-1, -9);
		\draw (-5, -9)[draw = black, fill = blue!50] rectangle 
		node{} (-4, -10);
		\draw (-4, -9)[draw = black, fill = blue!50] rectangle 
		node{} (-3, -10);
		\draw (-3, -9)[draw = black, fill = blue!50] rectangle 
		node{} (-2, -10);
		\draw (-2, -9)[draw = black, fill = blue!50] rectangle 
		node{} (-1, -10);
		\draw (-5, -10) rectangle node{$\dot{x}_{31}$} (-4, -11);
		\draw (-4, -10) rectangle node{$\dot{x}_{32}$} (-3, -11);
		\draw (-3, -10)[draw = black, fill = blue!50] rectangle 
		node{} (-2, -11);
		\draw (-2, -10)[draw = black, fill = blue!50] rectangle 
		node{} (-1, -11);
		\draw (-5, -11)[draw = black, fill = blue!50] rectangle 
		node{} (-4, -12);
		\draw (-4, -11)[draw = black, fill = blue!50] rectangle 
		node{} (-3, -12);
		\draw (-3, -11) rectangle node{$\dot{x}_{43}$} (-2, -12);
		\draw (-2, -11)[draw = black, fill = blue!50] rectangle 
		node{} (-1, -12);
		
		\draw (5, -8) rectangle node{$1$} (6, -9);
		\draw (6, -8) rectangle node{$0.5$} (7, -9);
		\draw (7, -8) rectangle node{$1$} (8, -9);
		\draw (8, -8) rectangle node{$1$} (9, -9);
		\draw (5, -9) rectangle node{$1$} (6, -10);
		\draw (6, -9) rectangle node{$1$} (7, -10);
		\draw (7, -9) rectangle node{$1$} (8, -10);
		\draw (8, -9) rectangle node{$0.5$} (9, -10);
		\draw (5, -10) rectangle node{$0$} (6, -11);
		\draw (6, -10) rectangle node{$0$} (7, -11);
		\draw (7, -10) rectangle node{$1$} (8, -11);
		\draw (8, -10) rectangle node{$1$} (9, -11);
		\draw (5, -11) rectangle node{$0.5$} (6, -12);
		\draw (6, -11) rectangle node{$1$} (7, -12);
		\draw (7, -11) rectangle node{$0.5$} (8, -12);
		\draw (8, -11) rectangle node{$1$} (9, -12);
		
		\draw[decorate, very thick, 
		decoration={brace, amplitude = 12pt, mirror}] 
		(-2.5, -12.2) -- (6.5, -12.2) node[midway] {};
		
		\node[draw, rounded corners = 3pt, minimum width = 3cm, 
		minimum height = 1.5cm] at (2, -14) {\textbf{Discriminator}};
		
		\draw[-latex, line width = 2pt] (2, -15.2)--(2, -15.8) 
		node[midway, right, xshift = 7pt, font = \footnotesize] 
		{Estimated Mask};
		
		\draw (0, -16) rectangle node{$1$} (1, -17);
		\draw (1, -16) rectangle node{$p_{12}$} (2, -17);
		\draw (2, -16) rectangle node{$1$} (3, -17);
		\draw (3, -16) rectangle node{$1$} (4, -17);
		\draw (0, -17) rectangle node{$1$} (1, -18);
		\draw (1, -17) rectangle node{$1$} (2, -18);
		\draw (2, -17) rectangle node{$1$} (3, -18);
		\draw (3, -17) rectangle node{$p_{24}$} (4, -18);
		\draw (0, -18) rectangle node{$0$} (1, -19);
		\draw (1, -18) rectangle node{$0$} (2, -19);
		\draw (2, -18) rectangle node{$1$} (3, -19);
		\draw (3, -18) rectangle node{$1$} (4, -19);
		\draw (0, -19) rectangle node{$p_{41}$} (1, -20);
		\draw (1, -19) rectangle node{$1$} (2, -20);
		\draw (2, -19) rectangle node{$p_{43}$} (3, -20);
		\draw (3, -19) rectangle node{$1$} (4, -20);
	\end{tikzpicture}
	\caption{An example of the GAIN workflow for imputation: The 
	original dataset (observed data in blue and missing data in red) 
	is decomposed into observed values, initial imputations (noise), 
	and a mask matrix indicating missingness; The generator uses 
	these inputs to produce imputed data, from which a reconstruction 
	loss is computed; The imputed data, together with a hint matrix 
	derived from the true mask, are passed to the discriminator, 
	which outputs an estimated mask; The adversarial loss is 
	calculated by comparing this estimated mask to the true mask.}
	\label{fig:gan}
\end{figure}

When applied to missing data imputation, GANs are often adapted into 
a specialized architecture: Generative Adversarial Imputation 
Network~\citep[GAIN,][]{yoon2018gain}. In GAIN, the generator 
produces a complete dataset by filling in missing entries based on 
observed features, the missingness mask, and random noise used to 
initialize imputed values. The discriminator, in turn, seeks to 
distinguish true observed values from imputations, aided by a hint 
mechanism that reveals partial information about the missingness 
pattern. The hint matrix is primarily introduced to prevent the 
discriminator from trivially recovering the full mask. This 
adversarial interplay drives the generator to iteratively refine its 
outputs so that imputations become statistically indistinguishable 
from true observations. See Figure~\ref{fig:gan} for an outline of 
GAIN workflow. 

Training relies on two complementary loss functions. First, a 
reconstruction loss ensures consistency to the observed data. 
Specifically, the generator outputs values for all data, and the loss 
is computed using the true observed entries only (e.g., the blue 
cells of the imputation matrix in Figure~\ref{fig:gan}), with mean 
squared error for continuous variables or cross-entropy for 
categorical variables by convention. Second, an adversarial loss is 
defined for the discriminator, which uses a cross-entropy objective 
to compare its estimated mask against the true mask, but only at 
entries left ambiguous (e.g., the cells with $0.5$ of the hint matrix 
in Figure~\ref{fig:gan}) by the hint matrix. The final 
objective of GAIN is a combination of these two losses, balancing 
agreement with observed values against the realism of imputations. 
Training proceeds until improvements in imputation quality stabilize.

Like MLP, implementation of GAIN in \proglang{R} relies on Python 
backends~\citep{yoon2018gain}, typically through 
\pkg{tensorflow}~\citep{pkg:tensorflow} in combination with 
\pkg{keras3}~\citep{pkg:keras3}. Following standard preprocessing 
steps such as normalization and construction of a binary mask to 
indicate observed versus missing data, both the generator and 
discriminator are formulated as multilayer neural networks 
initialized with Xavier weights, employing ReLU activations in hidden 
layers and sigmoid activations at the output. Training proceeds 
iteratively with Adam to enhance stability, and upon convergence, the 
generator yields imputed datasets that are subsequently re-scaled and 
transformed back to their original units.

GAN-based imputation methods have shown promise in handling 
high-dimensional and complex data, as they can naturally capture 
nonlinear dependencies and multimodal 
distributions~\citep{dai2021multiple}. They are 
particularly attractive in biomedical settings where missingness may 
be non-random and relationships among biomarkers are 
intricate~\citep{chen2023deep}. Model evaluation often relies on 
artificially masked validation sets, with performance assessed using 
RMSE, MAE, or classification accuracy, along with comparisons against 
benchmarks such as MICE, kNN, or MLP-based imputations. Despite these 
advantages, GAIN and similar methods come with 
several limitations. Training can be unstable and highly sensitive to 
hyperparameter choices, and convergence is not always guaranteed. In 
addition, interpretability is limited, and computational costs are 
typically higher than those of classical or simpler ML-based 
approaches~\citep{saxena2021GAN}.

\subsection{Variational Autoencoder (VAE)}
\label{sec:vae}

Variational autoencoders~\citep[VAEs,][]{kingma2014autoencoding} are 
a class of deep generative models that combine the flexibility of 
neural networks with probabilistic inference to approximate complex 
data distributions. Structurally, a VAE consists of two neural 
networks: an encoder that maps observed (possibly 
incomplete) data into a lower-dimensional latent space and a decoder 
that reconstructs the full data from samples drawn from that latent 
space. Unlike other neural architectures such as MLPs and GANs, VAEs 
explicitly impose a probabilistic structure by assuming a prior 
distribution (commonly Gaussian) on the latent variables. Training is 
performed by maximizing the evidence lower 
bound~\citep[ELBO,][]{blei2017variational}, which can be 
decomposed into a reconstruction loss that evaluates how 
well the decoder can reproduce the original data (typically using MSE 
for continuous variables or cross-entropy for categorical variables 
as the loss function) and a Kullback-Leibler divergence that 
penalizes 
discrepancies between the latent distribution learned by the encoder 
and the chosen prior. The ELBO formulation builds on the framework of 
variational inference in modern statistics, and provides 
a tractable solution, as opposed to intractable marginal 
likelihood which requires integrating over all possible latent 
variable configurations~\citep{hoffman2013stochastic}. This approach 
enables efficient joint optimization of both encoder and decoder 
parameters.

When adapted for missing data imputation, VAEs operate by treating 
the observed entries as partial input to the encoder, while missing 
entries are initially filled with random values or placeholders. The 
decoder then reconstructs the complete dataset, generating 
imputations for the missing components by leveraging learned 
dependencies in the latent representation. Training iterates between 
encoding incomplete data, sampling latent variables, and decoding 
full reconstructions, thereby producing imputations that are 
statistically consistent with both the observed values and the latent 
prior.

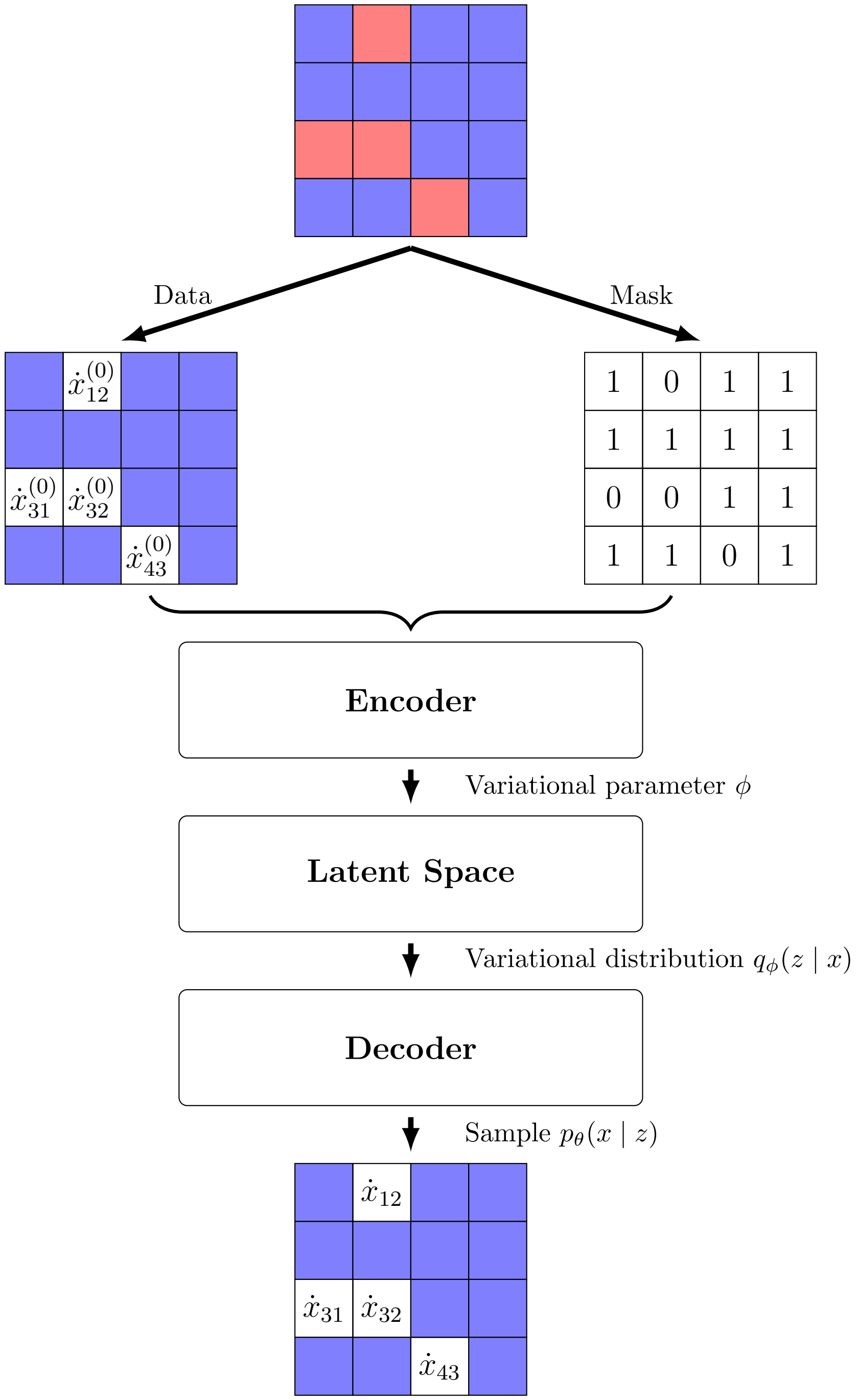
\begin{figure}
	\centering
	\begin{tikzpicture}[scale=0.75]
		\draw[draw = black, fill = blue!50] (0, 6) rectangle (1, 5);
		\draw[draw = black, fill = red!50] (1, 6) rectangle (2, 5);
		\draw[draw = black, fill = blue!50] (2, 6) rectangle (3, 5);
		\draw[draw = black, fill = blue!50] (3, 6) rectangle (4, 5);
		\draw[draw = black, fill = blue!50] (0, 5) rectangle (1, 4);
		\draw[draw = black, fill = blue!50] (1, 5) rectangle (2, 4);
		\draw[draw = black, fill = blue!50] (2, 5) rectangle (3, 4);
		\draw[draw = black, fill = blue!50] (3, 5) rectangle (4, 4);
		\draw[draw = black, fill = red!50] (0, 4) rectangle (1, 3);
		\draw[draw = black, fill = red!50] (1, 4) rectangle (2, 3);
		\draw[draw = black, fill = blue!50] (2, 4) rectangle (3, 3);
		\draw[draw = black, fill = blue!50] (3, 4) rectangle (4, 3);
		\draw[draw = black, fill = blue!50] (0, 3) rectangle (1, 2);
		\draw[draw = black, fill = blue!50] (1, 3) rectangle (2, 2);
		\draw[draw = black, fill = red!50] (2, 3) rectangle (3, 2);
		\draw[draw = black, fill = blue!50] (3, 3) rectangle (4, 2);
		
		\draw (5, 0) rectangle node{$1$} (6, -1);
		\draw (6, 0) rectangle node{$0$} (7, -1);
		\draw (7, 0) rectangle node{$1$} (8, -1);
		\draw (8, 0) rectangle node{$1$} (9, -1);
		\draw (5, -1) rectangle node{$1$} (6, -2);
		\draw (6, -1) rectangle node{$1$} (7, -2);
		\draw (7, -1) rectangle node{$1$} (8, -2);
		\draw (8, -1) rectangle node{$1$} (9, -2);
		\draw (5, -2) rectangle node{$0$} (6, -3);
		\draw (6, -2) rectangle node{$0$} (7, -3);
		\draw (7, -2) rectangle node{$1$} (8, -3);
		\draw (8, -2) rectangle node{$1$} (9, -3);
		\draw (5, -3) rectangle node{$1$} (6, -4);
		\draw (6, -3) rectangle node{$1$} (7, -4);
		\draw (7, -3) rectangle node{$0$} (8, -4);
		\draw (8, -3) rectangle node{$1$} (9, -4);
		
		\draw[draw = black, fill = blue!50] (-5, 0) rectangle 
		node{} (-4, -1);
		\draw (-4, 0) rectangle node{$\dot{x}_{12}^{(0)}$} 
		(-3, -1);
		\draw[draw = black, fill = blue!50] (-3, 0) rectangle 
		node{} (-2, -1);
		\draw[draw = black, fill = blue!50] (-2, 0) rectangle 
		node{} (-1, -1);
		\draw[draw = black, fill = blue!50] (-5, -1) rectangle 
		node{} (-4, -2);
		\draw[draw = black, fill = blue!50] (-4, -1) rectangle 
		node{} (-3, -2);
		\draw[draw = black, fill = blue!50] (-3, -1) rectangle 
		node{} (-2, -2);
		\draw[draw = black, fill = blue!50] (-2, -1) rectangle 
		node{} (-1, -2);
		\draw (-5, -2) rectangle node{$\dot{x}_{31}^{(0)}$} (-4, -3);
		\draw (-4, -2) rectangle node{$\dot{x}_{32}^{(0)}$} (-3, -3);
		\draw[draw = black, fill = blue!50] (-3, -2) rectangle 
		node{} (-2, -3);
		\draw[draw = black, fill = blue!50] (-2, -2) rectangle 
		node{} (-1, -3);
		\draw[draw = black, fill = blue!50] (-5, -3) rectangle 
		node{} (-4, -4);
		\draw[draw = black, fill = blue!50] (-4, -3) rectangle 
		node{} (-3, -4);
		\draw (-3, -3) rectangle node{$\dot{x}_{43}^{(0)}$} (-2, -4);
		\draw[draw = black, fill = blue!50] (-2, -3) rectangle 
		node{} (-1, -4);
		
		\draw[-latex, line width = 2pt] (2, 1.8)--(-3, 0.2)
		node[midway, left, xshift = -15pt, font = \footnotesize] 
		{Data};
		
		\draw[-latex, line width = 2pt] (2, 1.8)--(7, 0.2)
		node[midway, right, xshift = 15pt, font = \footnotesize] 
		{Mask};
		
		\draw[decorate, very thick, 
		decoration={brace, amplitude = 12pt, mirror}] 
		(-2.5, -4.2) -- (6.5, -4.2) node[midway] {};
		
		\node[draw, rounded corners = 3pt, minimum width = 6cm, 
		minimum height = 1.5cm] at (2, -6) {\textbf{Encoder}};
		
		\draw[-latex, line width = 2pt] (2, -7.2)--(2, -7.8)
		node[midway, right, xshift = 15pt, font = \footnotesize] 
		{Variational parameter $\phi$};
		
		\node[draw, rounded corners = 3pt, minimum width = 6cm, 
		minimum height = 1.5cm] at (2, -9) {\textbf{Latent Space}};
		
		\draw[-latex, line width = 2pt] (2, -10.2)--(2, -10.8)
		node[midway, right, xshift = 15pt, font = \footnotesize] 
		{Variational distribution $q_\phi(z \mid x)$};
		
		\node[draw, rounded corners = 3pt, minimum width = 6cm, 
		minimum height = 1.5cm] at (2, -12) {\textbf{Decoder}};
		
		\draw[-latex, line width = 2pt] (2, -13.2)--(2, -13.8) 
		node[midway, right, xshift = 15pt, font = \footnotesize] 
		{Sample $p_\theta(x \mid z)$};
		
		\draw (0, -14)[draw = black, fill = blue!50] rectangle 
		node{} (1, -15);
		\draw (1, -14) rectangle node{$\dot{x}_{12}$} (2, -15);
		\draw (2, -14) [draw = black, fill = blue!50] rectangle 
		node{} (3, -15);
		\draw (3, -14) [draw = black, fill = blue!50] rectangle 
		node{} (4, -15);
		\draw (0, -15) [draw = black, fill = blue!50] rectangle 
		node{} (1, -16);
		\draw (1, -15) [draw = black, fill = blue!50] rectangle 
		node{} (2, -16);
		\draw (2, -15) [draw = black, fill = blue!50] rectangle 
		node{} (3, -16);
		\draw (3, -15) [draw = black, fill = blue!50] rectangle 
		node{} (4, -16);
		\draw (0, -16) rectangle node{$\dot{x}_{31}$} (1, -17);
		\draw (1, -16) rectangle node{$\dot{x}_{32}$} (2, -17);
		\draw (2, -16) [draw = black, fill = blue!50] rectangle (3, 
		-17);
		\draw (3, -16) [draw = black, fill = blue!50] rectangle (4, 
		-17);
		\draw (0, -17) [draw = black, fill = blue!50] rectangle (1, 
		-18);
		\draw (1, -17) [draw = black, fill = blue!50] rectangle (2, 
		-18);
		\draw (2, -17) rectangle node{$\dot{x}_{43}$} (3, -18);
		\draw (3, -17) [draw = black, fill = blue!50] rectangle (4, 
		-18);
	\end{tikzpicture}
	\caption{An example of the VAE workflow for missing-data 
	imputation. The original dataset (observed entries in blue and 
	missing entries in red) is first represented as partially 
	observed values together with initial imputations (denoted by 
	superscript $(0)$) and a binary mask indicating missingness; The 
	encoder network takes these inputs and outputs the variational 
	parameters ($\phi$) that define the approximate posterior 
	$q_\phi(z\mid x)$; Latent variables $z$ are sampled from the 
	variational distribution and passed through the decoder 
	parameterized by $\theta$, which reconstructs the complete data 
	by producing imputed values (denoted by $\dot{x}$) for the 
	missing entries.}
	\label{fig:vae}
\end{figure}

Implementation of VAEs for imputation in \proglang{R} generally 
relies on the \pkg{tensorflow}~\citep{pkg:tensorflow} backend, which 
is accessed through \pkg{keras3}~\citep{pkg:keras3}. Similar to GAIN, 
preprocessing involves normalization and mask construction to 
indicate missingness. The encoder and decoder are typically 
parameterized as multilayer neural networks with ReLU activations in 
hidden layers and sigmoid outputs. Optimization is 
carried out with stochastic gradient descent or adaptive optimizers 
like Adam. After training, the decoder is applied iteratively to 
generate imputed datasets, which are then re-scaled to their original 
scale for downstream analysis. 

VAEs offer advantages in modeling uncertainty, as imputations can be 
drawn multiple times from the latent posterior to reflect 
variability, providing a natural framework for multiple imputation in 
practice~\citep{lall2022MIDAS}. However, VAE has several limitations 
when used for missing data imputation. First, like other deep 
learning algorithms, VAEs require careful architectural and 
hyperparameter specification, including latent space dimension and 
activation functions, which can substantially affect imputation 
quality and stability; besides, their performance often depends on 
large sample sizes~\citep{sun2023deep}. Second, because VAEs use a 
Gaussian prior for the latent space, misspecification of this prior 
may restrict its ability to capture complex, multimodal data 
features. Third, although VAEs naturally provide a probabilistic 
framework for uncertainty quantification, posterior samples may 
underestimate true variability, especially under MNAR 
mechanisms~\citep{roskams2023leveraging}. 
Finally, VAE training can be computationally demanding 
and sensitive to convergence issues, particularly when imputing 
high-dimensional data with substantial missingness. Interpretability 
is a persistent challenge across deep learning algorithms, and this 
is particularly true for VAEs, where the learned latent space often 
lacks a clear connection to clinically meaningful constructs.

\section{Simulations}
\label{sec:sim}

We conducted an extensive simulation study to evaluate the 
performance of six ML- and DL-based imputation methods described in 
Section~\ref{sec:method}; namely, they are $k$-nearest neighbors 
(kNN), support vector machines (SVM), random forests (RF), 
multilayer perceptrons (MLP), generative adversarial networks (GAN), 
and variational autoencoders (VAE). The performance of each method 
was assessed under a range of missing data scenarios that varied 
both in the mechanism (i.e., MCAR, MAR and MNAR) and in the proportion 
of missingness. For benchmarking, two traditional approaches, 
complete case analysis (CCA) and multiple imputation by chained 
equations~\citep[MICE,][]{vanbuuren2011mice}, were also included to 
provide a comprehensive comparison with these modern imputation 
methods.

\subsection{Synthetic Data Generation}
\label{sec:datagen}

The synthetic data generation was guided by the real data described 
in the next section (i.e., Section~\ref{sec:app}). The sample size 
was fixed at $n = 300$. The predictor subject to missingness mimicked 
the ratio of CSF A$\beta_{42}$ to CSF A$\beta_{40}$, generated from a 
Gamma distribution with shape parameter $9$ and scale parameter 
$0.008$, 
i.e., $\mathrm{CSF\ ratio} \sim \mathrm{Gamma}(9, 0.008)$. These 
parameter values were estimated via the method of moments using the 
real CSF data. The simulated CSF ratio values were standardized and 
denoted as $z_{\mathrm{csf}}$. These standardized values were not 
directly used in the regression model, but were retained for 
generating the correlated age variable subsequently.

Two additional covariates, age and sex, were considered. To induce a 
moderate correlation ($\rho = 0.6$) between CSF ratio and age, a 
standardized age variable $z_{\mathrm{age}}$ was generated using a 
Gaussian copula conditional on $z_{\mathrm{csf}}$. The final age 
variable was then defined as $\text{age} = 72 + 7 \times 
z_{\mathrm{age}}$, with the mean and standard deviation chosen to 
match those observed in the real data. Sex was generated 
independently from a Bernoulli distribution with probability $0.6$ 
for male.

To introduce additional right skewness and mild outliers, $10\%$ of 
the CSF ratio values were randomly selected and replaced with extreme 
values sampled from the $75$th to $95$th percentiles of the simulated 
CSF ratio distribution. Finally, the neuropsychological outcome 
variable was generated from a linear regression model,
\[
\text{neuropsych} = \beta_0 + \beta_1 \cdot \text{csfratio} + \beta_2 
\cdot \text{age} + \beta_3 \cdot \text{sex} + \varepsilon,
\]
where the true coefficients were set to $\beta_0 = 30$, $\beta_1 = 
18$, $\beta_2 = -0.2$, and $\beta_3 = -1.2$. The random error term 
$\varepsilon$ was sampled from a normal distribution with mean $0$ 
and standard deviation $1.5$, i.e., $\varepsilon \sim \mathcal{N}(0, 
1.5)$.

Missingness was then introduced into the CSF ratio variable under 
nine scenarios, defined by the combination of three missing data 
mechanisms (MCAR, MAR, and MNAR) and three missing proportions 
($\rho_{\text{miss}} = 25\%$, $50\%$, and $75\%$). In the MCAR 
setting, missing values were generated completely at random through 
simple random sampling. In the MAR setting, the probability of 
missingness in the CSF ratio depended on the two fully observed 
covariates, age and sex. In the MNAR setting, the probability of 
missingness depended on the CSF ratio itself, introducing 
non-ignorable missingness. 

To achieve a pre-specified missing proportion $\rho_{\text{miss}}$ 
under MAR and MNAR, we modeled the missingness indicator for each 
subject $i$ using Bernoulli with probability given by 
$\text{logit}^{-1}(\gamma_0 + \bm{X}_i^{\top}\bm{\gamma})$, where 
$\bm{X}_i$ contains the covariates used for the missing data 
generation (age and sex for MAR; csfratio for MNAR). We calibrated 
$(\gamma_0,\bm{\gamma})$ so that the squared error between expected 
missing proportion and $\rho_{\text{miss}}$ would be minimized. With 
the calibrated parameters $(\hat{\gamma}_0,\hat{\bm{\gamma}})$, we 
generated missingness independently for each subject. While this 
procedure does not guarantee that every simulated dataset 
attains $\rho_{\text{miss}}$ exactly, the actual proportion 
concentrates around the target, and the average deviation is small 
(typically less than $1\%$).

\subsection{Implementation and Evaluation}
\label{sec:impl}

The eight imputation methods were implemented in the \proglang{R} 
environment using a combination of existing \proglang{R} packages and 
adapted \proglang{Python} scripts, as summarized in 
Table~\ref{tab:implement_summary}. Conventional statistical and 
machine learning methods (e.g., MICE and RF) were implemented through 
well-established \proglang{R} packages, whereas deep learning 
methods, including MLP, VAE, and GAN, lack dedicated \proglang{R} 
packages and were therefore implemented via the \pkg{keras} and 
\pkg{tensorflow} interfaces within \proglang{R}. Details of the 
function specifications are provided in the online supplement, and 
all implementation codes are available in \url{https://github.com/paayq/ImputDL} to ensure reproducibility.

\begin{table}[tbp]
	\centering
	\renewcommand{\arraystretch}{1.25}
	\setlength{\tabcolsep}{13.6pt}
	\begin{tabular}{ll}
		\toprule
		Method & Source of Implementation \\ 
		\midrule
		CCA & Simple random sampling \\
		MICE & \pkg{mice}~\citep{vanbuuren2011mice} \\
		kNN & \pkg{VIM}~\citep{pkg:VIM} \\ 
		SVM & \pkg{e1071}~\citep{pkg:e1071} \\
		RF & \pkg{mice}~\citep{vanbuuren2011mice} \\ 
		MLP/GAN/VAE & \pkg{keras}~\citep{pkg:keras} \\ 
		& \pkg{tensorflow}~\citep{pkg:tensorflow} \\
		\bottomrule
	\end{tabular}
	\caption{Sources of implementation for eight imputation methods.}
	\label{tab:implement_summary}
\end{table}

For each simulation scenario, a total of $500$ replicates were 
conducted. Method performance was evaluated using four metrics: 
percentage bias (PB), empirical standard deviation (SD), standard 
error (SE) and coverage probability (CP). The empirical standard 
deviation was calculated across simulation replicates, while the 
average standard error was obtained from the fitted regression 
models. Comparing these two quantities, together with the coverage 
probability of nominal 95\% confidence intervals, allowed us to 
assess the reliability of statistical inference following imputation. 
In addition, the average runtime for each method was recorded to 
assess computational efficiency.

\subsection{Simulation Results}
\label{sec:simres}

\begin{table}[tbp]
	\centering
	\renewcommand{\arraystretch}{1.25}
	\small 
	\resizebox{\textwidth}{!}{%
	\setlength{\tabcolsep}{6.9pt}
	\begin{tabular}{l cccc cccc cccc}
		\toprule
		\multirow{2}{*}{} & 
		\multicolumn{4}{c}{$\rho_{\text{miss}} = 0.25$} & 
		\multicolumn{4}{c}{$\rho_{\text{miss}} = 0.50$} & 
		\multicolumn{4}{c}{$\rho_{\text{miss}} = 0.75$} \\
		\cmidrule(lr){2-5} \cmidrule(lr){6-9} \cmidrule(lr){10-13}
		& $\beta_0$ & $\beta_1$ & $\beta_2$ & $\beta_3$ & $\beta_0$ 
		& $\beta_1$ & $\beta_2$ & $\beta_3$ & $\beta_0$ & $\beta_1$ 
		& $\beta_2$ & $\beta_3$ \\ 
		\midrule
		CCA(PB) & 0.14 & 0.69 & 0.33 & 0.24 & 0.17 & 1.22 & 0.22 
		& 0.66 & 0.10 & 4.23 & 0.77 & 2.28 \\ 
		CCA(SD) & 1.12 & 4.05 & 0.01 & 0.20 & 1.32 & 5.17 & 0.02 & 
  		0.27 & 1.96 & 7.24 & 0.03 & 0.38 \\ 
  		CCA(SE) & 1.09 & 4.11 & 0.01 & 0.21 & 1.35 & 5.08 & 0.02 & 
  		0.25 & 1.94 & 7.32 & 0.03 & 0.36 \\ 
  		CCA(PC) & 0.95 & 0.94 & 0.95 & 0.95 & 0.94 & 0.95 & 0.95 & 
  		0.95 & 0.95 & 0.95 & 0.94 & 0.96 \\ 
  		\midrule
  		MICE(PB) & 0.01 & 0.96 & 0.01 & 0.83 & 0.27 & 2.78 & 0.30 & 
  		0.75 & 0.23 & 7.79 & 0.33 & 0.63 \\ 
  		MICE(SD) & 0.98 & 3.96 & 0.01 & 0.18 & 1.00 & 5.00 & 0.01 & 
  		0.19 & 1.14 & 6.82 & 0.01 & 0.20 \\ 
  		MICE(SE) & 0.97 & 4.04 & 0.01 & 0.18 & 1.01 & 4.88 & 0.01 & 
  		0.19 & 1.16 & 6.82 & 0.01 & 0.20 \\ 
  		MICE(PC) & 0.96 & 0.95 & 0.96 & 0.96 & 0.95 & 0.94 & 0.94 & 
  		0.95 & 0.97 & 0.95 & 0.96 & 0.96 \\ 
  		\midrule
  		kNN(PB) & 0.72 & 13.18 & 0.26 & 0.75 & 1.51 & 31.90 & 0.25 & 
  		0.73 & 2.88 & 45.56 & 1.96 & 0.13 \\ 
  		kNN(SD) & 1.03 & 4.62 & 0.01 & 0.18 & 1.15 & 7.12 & 0.01 & 
  		0.21 & 1.70 & 13.03 & 0.02 & 0.26 \\ 
  		kNN(SE) & 0.95 & 3.88 & 0.01 & 0.18 & 0.96 & 4.36 & 0.01 & 
  		0.18 & 1.00 & 5.31 & 0.01 & 0.18 \\ 
  		kNN(PC) & 0.96 & 0.93 & 0.96 & 0.95 & 0.94 & 0.90 & 0.95 & 
  		0.95 & 0.90 & 0.91 & 0.94 & 0.94 \\ 
  		\midrule
  		SVM(PB) & 1.13 & 23.92 & 0.11 & 0.77 & 2.66 & 61.59 & 0.09 & 
  		1.18 & 5.45 & 114.07 & 1.01 & 0.13 \\ 
  		SVM(SD) & 1.03 & 4.81 & 0.01 & 0.18 & 1.19 & 8.14 & 0.01 & 
  		0.21 & 1.95 & 16.47 & 0.02 & 0.31 \\ 
  		SVM(SE) & 0.94 & 3.98 & 0.01 & 0.18 & 0.95 & 4.65 & 0.01 & 
  		0.17 & 0.99 & 5.95 & 0.01 & 0.17 \\ 
  		SVM(PC) & 0.94 & 0.86 & 0.96 & 0.96 & 0.89 & 0.70 & 0.95 & 
  		0.94 & 0.85 & 0.74 & 0.94 & 0.95 \\ 
  		\midrule
  		RF(PB) & 0.83 & 20.67 & 0.15 & 0.95 & 1.89 & 41.46 & 0.06 & 
  		0.71 & 2.52 & 61.47 & 0.58 & 0.63 \\ 
  		RF(SD) & 0.96 & 3.25 & 0.01 & 0.18 & 0.95 & 3.14 & 0.01 & 
  		0.19 & 0.94 & 3.09 & 0.01 & 0.18 \\ 
  		RF(SE) & 0.98 & 4.12 & 0.01 & 0.18 & 1.01 & 4.64 & 0.01 & 
  		0.18 & 1.03 & 5.15 & 0.01 & 0.19 \\ 
  		RF(PC) & 0.94 & 0.81 & 0.96 & 0.96 & 0.91 & 0.33 & 0.95 & 
  		0.94 & 0.86 & 0.04 & 0.95 & 0.96 \\ 
  		\midrule
  		MLP(PB) & 0.01 & 0.68 & 0.01 & 0.97 & 0.20 & 1.30 & 0.30 & 
  		0.63 & 0.14 & 4.77 & 0.24 & 0.89 \\ 
  		MLP(SD) & 0.98 & 4.04 & 0.01 & 0.18 & 0.99 & 5.13 & 0.01 & 
  		0.19 & 1.04 & 7.11 & 0.01 & 0.18 \\ 
  		MLP(SE) & 0.96 & 4.15 & 0.01 & 0.18 & 1.00 & 5.17 & 0.01 & 
  		0.18 & 1.09 & 7.46 & 0.01 & 0.18 \\ 
  		MLP(PC) & 0.96 & 0.94 & 0.95 & 0.96 & 0.95 & 0.95 & 0.95 & 
  		0.95 & 0.96 & 0.95 & 0.96 & 0.96 \\ 
  		\midrule
  		GAN(PB) & 0.19 & 1.03 & 0.54 & 1.26 & 0.65 & 3.02 & 1.27 & 
  		0.79 & 2.96 & 9.80 & 5.51 & 1.94 \\ 
  		GAN(SD) & 1.09 & 5.19 & 0.01 & 0.18 & 1.42 & 9.77 & 0.01 & 
  		0.21 & 2.80 & 23.00 & 0.03 & 0.30 \\ 
  		GAN(SE) & 0.96 & 3.97 & 0.01 & 0.18 & 1.00 & 4.72 & 0.01 & 
  		0.18 & 1.09 & 6.45 & 0.01 & 0.19 \\ 
  		GAN(PC) & 0.95 & 0.95 & 0.95 & 0.96 & 0.95 & 0.95 & 0.95 & 
  		0.94 & 0.90 & 0.95 & 0.90 & 0.94 \\
  		\midrule
  		VAE(PB) & 0.08 & 0.43 & 0.16 & 0.85 & 0.06 & 0.77 & 0.05 & 
  		0.56 & 0.05 & 3.11 & 0.49 & 0.93 \\ 
  		VAE(SD) & 0.99 & 4.06 & 0.01 & 0.18 & 1.01 & 5.29 & 0.01 & 
  		0.19 & 1.06 & 7.31 & 0.01 & 0.18 \\ 
  		VAE(SE) & 0.96 & 4.12 & 0.01 & 0.18 & 1.00 & 5.12 & 0.01 & 
  		0.18 & 1.09 & 7.39 & 0.01 & 0.18 \\ 
  		VAE(PC) & 0.97 & 0.94 & 0.96 & 0.96 & 0.95 & 0.95 & 0.94 & 
  		0.95 & 0.96 & 0.95 & 0.96 & 0.96 \\  
		\bottomrule
	\end{tabular}}
  \caption{Simulation results for MCAR under various missing 
  proportions.}
  \label{tab:mcar}
\end{table}

Table~\ref{tab:mcar} summarizes the results under the MCAR mechanism 
across varying missing proportions. Both CCA and MICE provide 
unbiased estimates overall. As the missingness proportion 
$\rho_{\text{miss}}$ increases to 75\%, the bias in $\beta_1$ becomes 
slightly more notable for both methods: minimal for CCA and modestly 
higher for MICE. MICE, however, shows greater estimation efficiency 
than CCA, as reflected in smaller empirical standard deviations. 
Among the DL-based methods, MLP, GAN, and VAE also perform well under 
MCAR. GAN exhibits a minor bias in $\beta_1$, which becomes more 
noticeable as $\rho_{\text{miss}}$ increases. Moreover, GAN exhibits 
a 
substantial discrepancy between the empirical standard deviation and 
average estimated standard error for $\beta_1$, suggesting inadequate 
uncertainty calibration under the implementation considered. On the 
other hand, MLP and VAE maintain 
stable performance with comparable efficiency, both outperforming CCA
and achieving efficiency levels slightly lower than that of MICE. Among MICE, MLP, and 
VAE, the VAE exhibits the smallest bias in $\beta_1$. In contrast, 
the ML-based methods, kNN, SVM, and RF, show clear estimation bias 
for $\beta_1$, even under MCAR. Considering both bias and efficiency, 
VAE provides the most reliable performance in this setting.

\begin{table}[tbp]
	\centering
	\renewcommand{\arraystretch}{1.25}
	\small 
	\resizebox{\textwidth}{!}{
	\setlength{\tabcolsep}{6.9pt}
	\begin{tabular}{l cccc cccc cccc}
		\toprule
		\multirow{2}{*}{} & \multicolumn{4}{c}{$\rho_{\text{miss}} = 
		0.25$} & \multicolumn{4}{c}{$\rho_{\text{miss}} = 0.50$} & 
		\multicolumn{4}{c}{$\rho_{\text{miss}} = 0.75$} \\
		\cmidrule(lr){2-5} \cmidrule(lr){6-9} \cmidrule(lr){10-13}
		& $\beta_0$ & $\beta_1$ & $\beta_2$ & $\beta_3$ & $\beta_0$ 
		& $\beta_1$ & $\beta_2$ & $\beta_3$ & $\beta_0$ & $\beta_1$ 
		& $\beta_2$ & $\beta_3$ \\ 
		\midrule
		CCA(PB) & 0.06 & 1.25 & 0.17 & 0.49 & 0.08 & 2.87 & 0.12 & 
		0.48 & 0.21 & 2.02 & 0.14 & 1.26 \\ 
		CCA(SD) & 1.12 & 4.17 & 0.01 & 0.21 & 1.36 & 5.06 & 0.02 & 
		0.26 & 2.01 & 7.23 & 0.03 & 0.37 \\ 
		CCA(SE) & 1.09 & 4.12 & 0.01 & 0.20 & 1.35 & 5.09 & 0.02 & 
		0.25 & 1.94 & 7.35 & 0.03 & 0.36 \\ 
		CCA(PC) & 0.96 & 0.95 & 0.95 & 0.96 & 0.96 & 0.95 & 0.96 & 
		0.95 & 0.95 & 0.95 & 0.95 & 0.95 \\ 
		\midrule
		MICE(PB) & 0.03 & 1.35 & 0.00 & 0.94 & 0.23 & 3.81 & 0.13 & 
		0.68 & 0.29 & 5.69 & 0.02 & 0.65 \\ 
		MICE(SD) & 0.99 & 4.14 & 0.01 & 0.18 & 1.00 & 4.91 & 0.01 & 
		0.19 & 1.10 & 6.67 & 0.01 & 0.20 \\ 
		MICE(SE) & 0.97 & 4.05 & 0.01 & 0.18 & 1.02 & 4.89 & 0.01 & 
		0.19 & 1.15 & 6.83 & 0.01 & 0.20 \\ 
		MICE(PC) & 0.97 & 0.96 & 0.95 & 0.96 & 0.94 & 0.94 & 0.95 & 
		0.94 & 0.94 & 0.94 & 0.94 & 0.95 \\ 
  		\midrule
  		kNN(PB) & 0.69 & 13.38 & 0.17 & 0.88 & 1.53 & 29.96 & 0.48 & 
  		0.58 & 2.97 & 50.49 & 1.72 & 0.38 \\ 
  		kNN(SD) & 1.01 & 4.80 & 0.01 & 0.19 & 1.12 & 6.90 & 0.01 & 
  		0.20 & 1.59 & 12.13 & 0.02 & 0.26 \\ 
  		kNN(SE) & 0.95 & 3.89 & 0.01 & 0.18 & 0.96 & 4.37 & 0.01 & 
  		0.18 & 1.00 & 5.31 & 0.01 & 0.18 \\ 
  		kNN(PC) & 0.95 & 0.92 & 0.96 & 0.96 & 0.92 & 0.87 & 0.95 & 
  		0.95 & 0.91 & 0.90 & 0.94 & 0.95 \\ 
  		\midrule
  		SVM(PB) & 1.07 & 23.66 & 0.01 & 0.90 & 2.77 & 60.23 & 0.29 & 
  		0.69 & 5.77 & 122.43 & 0.88 & 0.58 \\ 
  		SVM(SD) & 1.02 & 5.10 & 0.01 & 0.19 & 1.18 & 7.85 & 0.01 & 
  		0.21 & 1.78 & 15.06 & 0.02 & 0.31 \\ 
  		SVM(SE) & 0.94 & 3.99 & 0.01 & 0.18 & 0.96 & 4.66 & 0.01 & 
  		0.17 & 0.99 & 5.94 & 0.01 & 0.17 \\ 
  		SVM(PC) & 0.94 & 0.87 & 0.95 & 0.95 & 0.89 & 0.70 & 0.96 & 
  		0.95 & 0.85 & 0.69 & 0.96 & 0.95 \\ 
  		\midrule
  		RF(PB) & 0.85 & 21.26 & 0.15 & 1.07 & 1.88 & 41.87 & 0.00 & 
  		0.62 & 2.50 & 60.34 & 0.50 & 0.71 \\ 
  		RF(SD) & 0.97 & 3.38 & 0.01 & 0.18 & 0.96 & 3.13 & 0.01 & 
  		0.19 & 0.95 & 2.98 & 0.01 & 0.18 \\ 
  		RF(SE) & 0.98 & 4.14 & 0.01 & 0.18 & 1.01 & 4.65 & 0.01 & 
  		0.18 & 1.03 & 5.17 & 0.01 & 0.19 \\ 
  		RF(PC) & 0.94 & 0.80 & 0.95 & 0.96 & 0.91 & 0.31 & 0.95 & 
  		0.95 & 0.86 & 0.05 & 0.95 & 0.96 \\ 
  		\midrule
  		MLP(PB) & 0.04 & 1.36 & 0.01 & 1.05 & 0.24 & 2.87 & 0.23 & 
  		0.54 & 0.02 & 1.83 & 0.19 & 0.94 \\ 
  		MLP(SD) & 0.98 & 4.16 & 0.01 & 0.18 & 0.99 & 5.02 & 0.01 & 
  		0.19 & 1.08 & 7.17 & 0.01 & 0.18 \\ 
  		MLP(SE) & 0.96 & 4.16 & 0.01 & 0.18 & 1.00 & 5.18 & 0.01 & 
  		0.18 & 1.09 & 7.49 & 0.01 & 0.18 \\ 
  		MLP(PC) & 0.96 & 0.95 & 0.96 & 0.95 & 0.94 & 0.96 & 0.95 & 
  		0.95 & 0.95 & 0.94 & 0.95 & 0.96 \\ 
  		\midrule
  		GAN(PB) & 0.26 & 1.99 & 0.78 & 0.84 & 1.08 & 5.86 & 1.92 & 
  		0.66 & 3.33 & 23.37 & 5.13 & 0.76 \\ 
  		GAN(SD) & 1.11 & 5.62 & 0.01 & 0.18 & 1.54 & 9.85 & 0.02 & 
  		0.20 & 3.01 & 23.47 & 0.03 & 0.28 \\ 
  		GAN(SE) & 0.96 & 3.96 & 0.01 & 0.18 & 1.00 & 4.72 & 0.01 & 
  		0.18 & 1.09 & 6.45 & 0.01 & 0.18 \\ 
  		GAN(PC) & 0.95 & 0.95 & 0.96 & 0.96 & 0.92 & 0.95 & 0.95 & 
  		0.94 & 0.92 & 0.93 & 0.92 & 0.94 \\ 
  		\midrule
  		VAE(PB) & 0.05 & 1.09 & 0.16 & 0.92 & 0.10 & 2.29 & 0.01 & 
  		0.49 & 0.18 & 0.08 & 0.45 & 0.98 \\ 
  		VAE(SD) & 0.99 & 4.15 & 0.01 & 0.18 & 1.00 & 5.12 & 0.01 & 
  		0.19 & 1.10 & 7.37 & 0.01 & 0.18 \\ 
  		VAE(SE) & 0.96 & 4.13 & 0.01 & 0.18 & 1.01 & 5.13 & 0.01 & 
  		0.18 & 1.09 & 7.42 & 0.01 & 0.18 \\ 
 		VAE(PC) & 0.96 & 0.94 & 0.95 & 0.96 & 0.95 & 0.95 & 0.95 & 
 		0.94 & 0.95 & 0.94 & 0.95 & 0.96 \\ 
  		\bottomrule
	\end{tabular}}
  \caption{Simulation results for MAR covariates under various 
  missing proportions.}
  \label{tab:mar}
\end{table}

Table~\ref{tab:mar} summarizes the results under the MAR mechanism 
across various levels of missingness. The findings are largely 
similar to those for the MCAR scenario. Both CCA and MICE yield 
unbiased estimates, while MICE presents higher efficiency than CCA. 
Among the DL-based methods, MLP and VAE produce remarkably accurate 
parameter estimates even when the missing proportion reaches as high 
as $75\%$, whereas GAN exhibits substantial bias in estimating 
$\beta_1$ when the missing proportion is high. MLP and VAE show 
comparable efficiency. They are slightly less efficient than MICE 
for estimating $\beta_1$, but more efficient for the other 
parameters. By contrast, the ML-based methods, including kNN, SVM 
and RF, once again fail to estimate $\beta_1$ accurately, showing 
significant bias. Overall, VAE delivers the best performance in this 
setting, with MLP remaining highly competitive. It is worth noting 
that we have carried out an additional set of simulations for the 
MAR scenario, in which missingness for ``csfratio'' depends not only 
on observed covariates but also on observed outcomes. The findings 
are extremely similar to those in Table~\ref{tab:mar}, and are 
therefore omitted here.

\begin{table}[tbp]
	\centering
	\renewcommand{\arraystretch}{1.25}
	\small 
	\resizebox{\textwidth}{!}{%
		\setlength{\tabcolsep}{6.9pt}
		\begin{tabular}{l cccc cccc cccc}
			\toprule
			\multirow{2}{*}{} & 
			\multicolumn{4}{c}{$\rho_{\text{miss}} = 0.25$} & 
			\multicolumn{4}{c}{$\rho_{\text{miss}} = 0.50$} & 
			\multicolumn{4}{c}{$\rho_{\text{miss}} = 0.75$} \\
			\cmidrule(lr){2-5} \cmidrule(lr){6-9} 
			\cmidrule(lr){10-13}
			& $\beta_0$ & $\beta_1$ & $\beta_2$ & $\beta_3$ & 
			$\beta_0$ & $\beta_1$ & $\beta_2$ & $\beta_3$ & 
			$\beta_0$ & $\beta_1$ & $\beta_2$ & $\beta_3$ \\ 
			\midrule
    		CCA(PB) & 0.03 & 1.51 & 0.15 & 0.45 & 0.16 & 0.22 & 0.39 
    		& 2.01 & 0.19 & 1.67 & 0.12 & 0.99 \\ 
    		CCA(SD) & 1.11 & 4.19 & 0.01 & 0.21 & 1.34 & 5.30 & 0.02 
    		& 0.26 & 2.00 & 7.22 & 0.03 & 0.37 \\ 
    		CCA(SE) & 1.09 & 4.12 & 0.01 & 0.20 & 1.35 & 5.09 & 0.02 
    		& 0.25 & 1.95 & 7.37 & 0.03 & 0.36 \\ 
    		CCA(PC) & 0.95 & 0.96 & 0.95 & 0.96 & 0.96 & 0.94 & 0.96 
    		& 0.95 & 0.95 & 0.95 & 0.96 & 0.95 \\ 
  			\midrule
  			MICE(PB) & 0.04 & 1.47 & 0.00 & 0.97 & 0.18 & 1.63 & 
  			0.27 & 1.17 & 0.30 & 5.66 & 0.04 & 0.53 \\ 
  			MICE(SD) & 0.99 & 4.17 & 0.01 & 0.18 & 1.01 & 5.12 & 
  			0.01 & 0.20 & 1.11 & 6.56 & 0.01 & 0.20 \\ 
  			MICE(SE) & 0.97 & 4.06 & 0.01 & 0.18 & 1.02 & 4.88 & 
  			0.01 & 0.19 & 1.16 & 6.83 & 0.01 & 0.20 \\ 
  			MICE(PC) & 0.96 & 0.95 & 0.95 & 0.95 & 0.95 & 0.94 & 
  			0.95 & 0.95 & 0.95 & 0.94 & 0.95 & 0.96 \\ 
  			\midrule
  			kNN(PB) & 0.70 & 13.22 & 0.20 & 0.94 & 1.61 & 33.82 & 
  			0.22 & 1.40 & 2.91 & 51.15 & 1.52 & 0.13 \\ 
  			kNN(SD) & 1.02 & 4.83 & 0.01 & 0.19 & 1.15 & 7.05 & 0.01 
  			& 0.21 & 1.59 & 11.97 & 0.02 & 0.26 \\ 
  			kNN(SE) & 0.95 & 3.89 & 0.01 & 0.18 & 0.96 & 4.36 & 0.01 
  			& 0.18 & 1.00 & 5.31 & 0.01 & 0.18 \\ 
  			kNN(PC) & 0.96 & 0.93 & 0.96 & 0.96 & 0.92 & 0.86 & 0.95 
  			& 0.95 & 0.92 & 0.89 & 0.94 & 0.94 \\ 
  			\midrule
  			SVM(PB) & 1.08 & 23.55 & 0.04 & 0.92 & 2.86 & 64.05 & 
  			0.05 & 1.58 & 5.78 & 123.62 & 0.78 & 1.23 \\ 
  			SVM(SD) & 1.03 & 5.12 & 0.01 & 0.19 & 1.23 & 8.24 & 0.01 
  			& 0.21 & 1.81 & 14.87 & 0.02 & 0.31 \\ 
  			SVM(SE) & 0.94 & 3.99 & 0.01 & 0.18 & 0.95 & 4.64 & 0.01 
  			& 0.17 & 0.99 & 5.95 & 0.01 & 0.17 \\ 
  			SVM(PC) & 0.94 & 0.88 & 0.95 & 0.96 & 0.88 & 0.71 & 0.95 
  			& 0.94 & 0.84 & 0.67 & 0.96 & 0.93 \\ 
  			\midrule
  			RF(PB) & 0.86 & 21.49 & 0.15 & 1.09 & 1.83 & 40.86 & 
  			0.02 & 0.93 & 2.46 & 59.70 & 0.52 & 0.64 \\ 
  			RF(SD) & 0.97 & 3.36 & 0.01 & 0.18 & 0.96 & 3.16 & 0.01 
  			& 0.19 & 0.95 & 2.97 & 0.01 & 0.18 \\ 
  			RF(SE) & 0.98 & 4.14 & 0.01 & 0.18 & 1.01 & 4.62 & 0.01 
  			& 0.18 & 1.03 & 5.16 & 0.01 & 0.19 \\ 
  			RF(PC) & 0.93 & 0.78 & 0.95 & 0.95 & 0.92 & 0.35 & 0.95 
  			& 0.95 & 0.87 & 0.04 & 0.96 & 0.96 \\ 
  			\midrule
  			MLP(PB) & 0.04 & 1.62 & 0.00 & 1.06 & 0.13 & 0.22 & 
  			0.27 & 0.79 & 0.03 & 1.59 & 0.18 & 0.92 \\ 
  			MLP(SD) & 0.98 & 4.18 & 0.01 & 0.18 & 1.02 & 5.28 & 0.01 
  			& 0.19 & 1.07 & 7.12 & 0.01 & 0.18 \\ 
  			MLP(SE) & 0.96 & 4.17 & 0.01 & 0.18 & 1.00 & 5.18 & 0.01 
  			& 0.18 & 1.09 & 7.50 & 0.01 & 0.18 \\ 
  			MLP(PC) & 0.96 & 0.95 & 0.95 & 0.95 & 0.95 & 0.95 & 0.95 
  			& 0.95 & 0.96 & 0.95 & 0.95 & 0.96 \\ 
  			\midrule
  			GAN(PB) & 0.13 & 3.48 & 0.76 & 0.06 & 1.17 & 5.52 & 
  			2.01 & 0.61 & 2.92 & 14.14 & 5.04 & 0.82 \\ 
  			GAN(SD) & 1.08 & 5.24 & 0.01 & 0.17 & 1.49 & 10.23 & 
  			0.02 & 0.20 & 2.86 & 24.56 & 0.03 & 0.29 \\ 
  			GAN(SE) & 0.96 & 3.97 & 0.01 & 0.18 & 1.00 & 4.78 & 0.01 
  			& 0.18 & 1.08 & 6.43 & 0.01 & 0.19 \\ 
  			GAN(PC) & 0.95 & 0.95 & 0.96 & 0.95 & 0.94 & 0.96 & 0.94 
  			& 0.96 & 0.91 & 0.95 & 0.92 & 0.95 \\
  			\midrule
  			VAE(PB) & 0.05 & 1.28 & 0.17 & 0.94 & 0.02 & 0.45 & 0.02 
  			& 0.70 & 0.17 & 0.11 & 0.43 & 0.96 \\ 
  			VAE(SD) & 0.99 & 4.18 & 0.01 & 0.18 & 1.03 & 5.34 & 0.01 
  			& 0.19 & 1.10 & 7.31 & 0.01 & 0.18 \\ 
  			VAE(SE) & 0.96 & 4.14 & 0.01 & 0.18 & 1.01 & 5.14 & 0.01 
  			& 0.18 & 1.09 & 7.43 & 0.01 & 0.18 \\ 
  			VAE(PC) & 0.96 & 0.95 & 0.95 & 0.95 & 0.94 & 0.95 & 0.95 
  			& 0.95 & 0.95 & 0.95 & 0.95 & 0.96 \\ 
			\bottomrule
		\end{tabular}}
	\caption{Simulation results for MNAR under various missing 
	proportions.}
	\label{tab:mnar}
\end{table}

Although the missing data mechanism is MNAR (i.e., CSF missingness 
depends on the CSF value itself), the results in Table~\ref{tab:mnar} 
are similar to those observed under MCAR and MAR. Overall, CCA, MICE, MLP, and VAE yield essentially unbiased estimates. VAE achieves the highest overall accuracy across all missingness proportions, whereas the performance of MICE degrades as the missingness proportion increases and becomes the worst among the methods at high levels of missingness. Nevertheless, MICE presents greater efficiency 
than CCA across all parameters and missingness levels. For $\beta_1$, 
MLP and VAE are slightly less efficient than MICE but tend to 
outperform it for other coefficients, although the efficiency 
differences among MICE, MLP, and VAE are generally small. GAN 
produces unbiased estimates when the missing proportion is below 
$50\%$, but the bias in $\beta_1$ becomes obvious at higher missing 
rates. MLP exhibits estimation inconsistency for $\beta_1$, which 
worsens as the missing proportion increases. Other methods, including 
kNN, SVM, and RF, perform poorly overall in estimating $\beta_1$. 
Among all methods, VAE appears most reliable, while MICE and MLP 
remain competitive alternatives.

The persistence of unbiasedness for CCA under this MNAR setting can 
be explained by the fact that, in a linear model, missingness 
depending solely on regressors (even on the unobserved covariate 
itself) does not affect the conditional expectation of the outcome, 
as the error maintains mean zero given the covariates. Consequently, 
CCA remains unbiased for slope parameters, though less efficient due 
to reduced sample size. However, this property does not necessarily 
extend to MNAR mechanisms in which missingness also depends on the 
outcome, as the conditional mean-zero assumption for the regression 
error may no longer hold among complete cases. Although MICE formally 
assumes MAR, when the imputation model is correctly specified and the 
relevant associations are adequately informed by the observed data, 
multiple imputation preserves the correct covariate-outcome 
relationships. As a result, MI yields unbiased regression 
coefficients and offers improved precision compared to CCA. Bias 
would arise, however, if missingness depended on the outcome 
variable. To confirm this, we conducted an additional set of 
MNAR-based simulation studies where missingness was also
directly associated with the outcome.

\subsection{Additional MNAR Experiments}
\label{sec:mnar_extra}

We conducted an additional set of MNAR experiments in which the 
missingness depended on both the unobserved covariate csfratio and the observed outcome. 
The data generation framework largely followed 
Section~\ref{sec:datagen}, with minor modifications designed to 
introduce greater variability and right-skewness in the covariate 
distribution. These adjustments were made to accentuate potential 
bias under MNAR, thereby highlighting the importance of using 
appropriate imputation strategies for MNAR. Specifically: (1) 
correlations among all covariates, including sex, were introduced 
through a Gaussian copula with $\rho = 0.6$; (2) the Gamma 
distribution used to generate CSF ratio values was adjusted to 
increase right‐tail heaviness; and (3) the error variance in the 
outcome model was increased to produce more heterogeneous outcomes.

Missing CSF ratio values were then generated under an MNAR mechanism 
as follows. First, the simulated CSF ratio values were normalized and 
transformed using a fourth‐power function to enhance right‐tail 
effects, creating $s_i$ for each subject $i$. The probability of CSF 
missingness for subject $i$ was defined as
\[
p_{\text{miss}, i} = \text{logit}^{-1}(-1 + 5s_i) + 0.3 \times 
\mathbf{1}_{\{\text{neuropsych} > 26 \text{ or } < 11\}} + 
\text{Unif}(-0.08, 0.08),
\]
where $\mathbf{1}_{\{\cdot\}}$ denotes the indicator function and 
$\text{Unif}(\cdot)$ denotes a continuous uniform random variable.
The first term ensures that missingness depends directly on the 
(unobserved) CSF variable itself, establishing MNAR. The indicator 
function increases the likelihood of missingness among individuals 
who are either cognitively healthy or severely impaired, reflecting 
realistic clinical patterns. The added uniform noise introduces 
additional variability in missingness rates across subjects.

Finally, to guarantee all subjects retain a nonzero probability of 
having their CSF observed, including extreme values crucial for 
unbiased learning, we truncated missingness probabilities to the 
interval $[0.05, 0.95]$:
\[p_{\text{miss}, i} \leftarrow \min\{0.95, \max\{0.05, 
p_{\text{miss}, i}\}\}.\]
This setup ensures that all methods, including deep learning 
approaches, can access informative regions of the data, supporting 
more stable imputation performance. The parameters governing the 
missingness model were calibrated to yield an overall missing 
proportion of approximately 58\%, consistent with the real data 
example in Section~\ref{sec:app}, while simultaneously creating a 
challenging scenario to rigorously assess the robustness of each 
method. Motivated by prior work~\citep[e.g.,][]{sun2017revisiting, 
karras2020training} which shows that deep learning methods (e.g., 
GANs) generally yield more reliable performance with larger training 
sets, we also evaluate 
multiple sample sizes, $n = \{300, 1{,}000, 2{,}000\}$, in this 
experiment.

\begin{table}[tbp]
	\centering
	\renewcommand{\arraystretch}{1.15}
	\small
	\resizebox{\textwidth}{!}{
		\setlength{\tabcolsep}{6.5pt}
		\begin{tabular}{l cccc cccc cccc}
			\toprule
			\multirow{2}{*}{} 
			& \multicolumn{4}{c}{$n = 300$} 
			& \multicolumn{4}{c}{$n = 1{,}000$}
			& \multicolumn{4}{c}{$n = 2{,}000$} \\
			\cmidrule(lr){2-5} \cmidrule(lr){6-9} \cmidrule(lr){10-13}
			& $\beta_0$ & $\beta_1$ & $\beta_2$ & $\beta_3$
			& $\beta_0$ & $\beta_1$ & $\beta_2$ & $\beta_3$
			& $\beta_0$ & $\beta_1$ & $\beta_2$ & $\beta_3$ \\
			\midrule
			CCA(PB) & 3.66 & 11.61 & 10.79 & 11.35 & 3.81 & 
			11.47 & 10.88 & 12.25 & 4.15 & 11.46 & 11.38 & 
			10.10 \\
			CCA(SD) & 3.69 & 2.69 & 0.05 & 0.66 & 2.02 & 2.01 & 0.03 
			& 0.37 & 1.53 & 1.66 & 0.02 & 0.26 \\
			CCA(SE) & 3.69 & 2.54 & 0.05 & 0.70 & 2.05 & 1.92 & 0.03 
			& 0.38 & 1.51 & 1.58 & 0.02 & 0.28 \\
			CCA(PC) & 0.94 & 0.89 & 0.92 & 0.93 & 0.92 & 0.83 & 0.78 
			& 0.92 & 0.88 & 0.77 & 0.61 & 0.91 \\
			\midrule
			MICE(PB) & 8.76 & 46.79 & 0.75 & 2.04 & 6.65 & 35.56 & 
			0.05 & 0.13 & 5.53 & 28.69 & 0.22 & 1.36 \\
			MICE(SD) & 3.94 & 3.50 & 0.05 & 0.71 & 2.09 & 2.52 & 0.03 
			& 0.37 & 1.44 & 2.07 & 0.02 & 0.24 \\
			MICE(SE) & 3.97 & 2.84 & 0.05 & 0.74 & 2.01 & 2.36 & 0.03 
			& 0.35 & 1.40 & 1.93 & 0.02 & 0.24 \\
			MICE(PC) & 0.91 & 0.33 & 0.94 & 0.95 & 0.85 & 0.27 & 0.95 
			& 0.94 & 0.79 & 0.28 & 0.94 & 0.95 \\
			\midrule
			kNN(PB) & 10.46 & 66.80 & 3.57 & 8.08 & 13.45 & 71.56 & 
			1.10 & 1.83 & 15.08 & 75.90 & 0.22 & 1.90 \\
			kNN(SD) & 4.77 & 5.20 & 0.06 & 0.88 & 2.51 & 3.95 & 0.03 
			& 0.42 & 1.81 & 3.33 & 0.02 & 0.27 \\
			kNN(SE) & 3.59 & 3.14 & 0.05 & 0.68 & 1.74 & 2.02 & 0.02 
			& 0.32 & 1.19 & 1.55 & 0.01 & 0.21 \\
			kNN(PC) & 0.90 & 0.37 & 0.94 & 0.94 & 0.64 & 0.10 & 0.95 
			& 0.95 & 0.32 & 0.01 & 0.95 & 0.95 \\
			\midrule
			SVM(PB) & 14.24 & 81.39 & 3.55 & 5.57 & 19.71 & 100.36 
			& 0.99 & 1.42 & 23.75 & 117.61 & 0.32 & 1.89 \\
			SVM(SD) & 4.66 & 4.99 & 0.06 & 0.86 & 2.67 & 4.40 & 0.03 
			& 0.45 & 2.01 & 3.94 & 0.02 & 0.30 \\
			SVM(SE) & 3.53 & 3.19 & 0.05 & 0.66 & 1.72 & 2.12 & 0.02 
			& 0.31 & 1.18 & 1.66 & 0.01 & 0.21 \\
			SVM(PC) & 0.85 & 0.14 & 0.94 & 0.95 & 0.40 & 0.01 & 0.94 
			& 0.95 & 0.06 & 0.00 & 0.96 & 0.95 \\
			\midrule
			RF(PB) & 12.92 & 38.69 & 0.83 & 5.62 & 11.49 & 46.54 & 
			0.34 & 1.64 & 11.27 & 50.28 & 0.14 & 2.06 \\
			RF(SD) & 3.71 & 1.92 & 0.05 & 0.75 & 1.87 & 1.29 & 0.03 & 
			0.36 & 1.25 & 1.02 & 0.02 & 0.23 \\
			RF(SE) & 4.20 & 4.05 & 0.05 & 0.77 & 1.98 & 2.52 & 0.02 & 
			0.35 & 1.35 & 1.91 & 0.02 & 0.23 \\
			RF(PC) & 0.82 & 0.05 & 0.96 & 0.95 & 0.53 & 0.00 & 0.94 & 
			0.94 & 0.24 & 0.00 & 0.95 & 0.96 \\
			\midrule
			MLP(PB) & 6.89 & 11.09 & 2.06 & 4.71 & 4.08 & 11.04 & 
			0.98 & 1.02 & 3.20 & 11.03 & 0.38 & 1.69 \\
			MLP(SD) & 3.75 & 2.69 & 0.05 & 0.76 & 1.95 & 2.02 & 0.03 
			& 0.36 & 1.35 & 1.67 & 0.02 & 0.24 \\
			MLP(SE) & 4.15 & 4.24 & 0.05 & 0.76 & 1.97 & 2.66 & 0.02 
			& 0.35 & 1.36 & 2.06 & 0.02 & 0.23 \\
			MLP(PC) & 0.90 & 0.91 & 0.95 & 0.95 & 0.90 & 0.84 & 0.94 
			& 0.94 & 0.90 & 0.78 & 0.96 & 0.96 \\
			\midrule
			GAN(PB) & 6.57 & 13.49 & 8.74 & 10.58 & 4.92 & 7.96 & 
			6.45 & 2.02 & 4.84 & 0.61 & 8.29 & 4.93 \\
			GAN(SD) & 9.39 & 14.88 & 0.09 & 0.95 & 6.93 & 12.04 & 
			0.06 & 0.52 & 6.88 & 12.75 & 0.05 & 0.44 \\
			GAN(SE) & 3.92 & 3.64 & 0.05 & 0.72 & 1.93 & 2.38 & 0.02 
			& 0.34 & 1.33 & 1.82 & 0.02 & 0.23 \\
			GAN(PC) & 0.94 & 0.94 & 0.94 & 0.94 & 0.92 & 0.93 & 0.92 
			& 0.95 & 0.93 & 0.94 & 0.90 & 0.95 \\
			\midrule
			VAE(PB) & 5.98 & 8.38 & 1.69 & 5.31 & 2.63 & 5.92 & 
			0.53 & 1.41 & 1.61 & 5.29 & 0.14 & 1.96 \\
			VAE(SD) & 3.81 & 2.91 & 0.05 & 0.75 & 1.99 & 2.19 & 0.03 
			& 0.36 & 1.39 & 1.81 & 0.02 & 0.23 \\
			VAE(SE) & 4.14 & 4.19 & 0.05 & 0.76 & 1.97 & 2.62 & 0.02 
			& 0.34 & 1.36 & 2.03 & 0.02 & 0.23 \\
			VAE(PC) & 0.90 & 0.93 & 0.95 & 0.95 & 0.93 & 0.92 & 0.95 
			& 0.95 & 0.94 & 0.90 & 0.95 & 0.96 \\
			\bottomrule
	\end{tabular}}
	\caption{Simulation results for MNAR with missingness depending 
	on both unobserved covariates and outcomes; The missing 
	proportion was fixed around $0.58$ across all scenarios.}
	\label{tab:mnar_extra}
\end{table}

As summarized in Table~\ref{tab:mnar_extra}, CCA produces biased 
estimates for all regression coefficients. For MICE, the estimate of 
$\beta_1$ is severely biased, whereas the remaining coefficients are 
estimated reasonably well. MLP shows a mild bias for $\beta_1$ (with 
the other coefficients appearing unbiased), substantially smaller 
than MICE. The performance of MLP is overall insensitive to 
increases in sample size. The parameter estimation bias for GAN 
decreases as sample size grows. However, the pronounced gap between 
SE and empirical SD persists for $\beta_1$, indicating a lack of 
estimation consistency despite unbiased point estimates and 
satisfactory nominal probability coverage. It seems that, with a 
sufficiently large training set, GAN becomes more competitive. Among 
the methods considered, VAE exhibits the best overall performance, 
with relatively small estimation bias and improved precision as the 
sample size increases. These findings suggest that VAE is a promising 
approach under the MNAR mechanism considered in this experiment. By 
contrast, the other three ML-based methods, kNN, SVM, and RF, 
continue to perform poorly under this MNAR setting. Together, these 
results suggest that VAE is the most reliable imputation approach in 
this scenario.

\subsection{Runtime Summary}
\label{sec:runtime}

Table~\ref{tab:runtime} reports the mean runtime across $500$ 
replicates for each method. Among MICE, MLP, GAN, and VAE, VAE is 
the fastest, followed by MICE; MLP and GAN have comparable runtimes 
due to their similar implementation in our code. Runtime is 
generally stable across missing data mechanisms (MCAR, MAR, MNAR) 
and missingness proportions. While runtime can depend on hardware 
and implementation details, these results support prioritizing VAE 
for imputation in MCAR, MAR, and especially MNAR settings, given its 
favorable combination of speed and statistical performance.

\begin{table}[tbp]
	\centering
	\renewcommand{\arraystretch}{1.25}
	\small
	\resizebox{\textwidth}{!}{%
		\setlength{\tabcolsep}{6.9pt}
		\begin{tabular}{c c cccccccc}
			\toprule
			& & \multicolumn{8}{c}{Missing Data Methods} \\
			\cmidrule{3-10}
			& $\rho_{\text{miss}}$ & CCA & MICE & kNN & SVM & RF	
			& MLP 
			& GAN & VAE \\
			\midrule
			\multirow{3}{*}{MCAR} & 0.25 & $<0.001$ & 1.90 & 0.016 & 
			0.004 & 1.81 & 3.65 & 3.65 & 0.410 \\
			& 0.50 & $<0.001$ & 1.88 & 0.025 & 0.003 & 2.08 & 3.44 & 
			3.57 &	0.402 \\
			& 0.75 & $<0.001$ & 2.04 & 0.030 & 0.002 & 1.91 & 3.85 & 
			3.57 & 0.432 \\
			\midrule
			\multirow{3}{*}{MAR} & 0.25 & $<0.001$ & 1.80 & 0.016 & 
			0.004 & 2.15 & 3.74 & 3.58 & 0.383 \\
			& 0.50 & $<0.001$ & 1.92 & 0.025 & 0.003 & 1.87 & 5.69 & 
			3.57 & 0.405 \\
			& 0.75 & $<0.001$ & 1.79 & 0.029 & 0.003 & 2.03 & 4.01 & 
			3.57 & 0.386 \\
			\midrule
			\multirow{3}{*}{MNAR} & 0.25 & $<0.001$ & 1.73 & 0.017 & 
			0.003 & 1.60 & 2.42 & 3.57 & 0.371 \\
			& 0.50 & $<0.001$ & 1.74 & 0.026 & 0.003 & 1.66 & 2.50 & 
			3.59 & 0.376 \\
			& 0.75 & $<0.001$ & 1.71 & 0.031 & 0.002 & 1.63 & 2.50 & 
			3.54 & 0.373 \\
			\bottomrule
		\end{tabular}
	}
	\caption{Average runtime (in seconds) for each missing data 
	method across various missing data scenarios.}
	\label{tab:runtime}
\end{table}

\section{Applications}
\label{sec:app}

In this section, we apply the missing data imputation methods evaluated in this 
paper to an AD cohort. We analyzed data from the Vanderbilt Memory 
\& Aging Project (VMAP), which is an observational study of older 
adults without dementia at entry~\citep{osborn2018cerebrospinal}. We 
used 
the September 2025 data freeze, and our primary predictor is the 
baseline CSF A$\beta_{42}$/A$\beta_{40}$ ratio, motivated by 
evidence linking this ratio to increased risk of cognitive decline 
in AD~\citep{delaby2022abetaratio}. The VMAP study aims, diagnostic 
criteria, methods, and phenotyping have been described 
previously~\citep{osborn2018cerebrospinal}. Our goal is to assess the 
association between the baseline CSF A$\beta_{42}$/A$\beta_{40}$ 
ratio and cognitive status.

VMAP includes legacy and expansion cohorts. Because CSF biomarkers 
for the expansion cohort have not yet been processed, we restricted 
analyses to the legacy cohort ($n = 335$), which comprises $40.7\%$ 
women and has a mean age of $72.9 \pm 7.3$ years. The outcome is the 
Montreal Cognitive Assessment (MoCA), a global cognitive measure 
ranging from $0$ to $30$, with higher scores indicating better 
cognition. Because we focus on missing covariates, one participant 
with missing MoCA has been removed, resulting in $335$ participants 
with complete age, sex, and MoCA data. The CSF ratio is missing for 
$53.8\%$ of the cohort.

To assess the missing data mechanism, we applied Little’s MCAR test, 
which yielded $\text{p} = 0.017$, indicating that CSF ratio is 
unlikely to be MCAR. We then fit a logistic regression for the 
missingness indicator, adjusting for age, sex, and MoCA, and found 
that missingness was significantly associated with sex ($\text{p} = 
0.007$) but not age ($\text{p} = 0.663$) or MoCA ($\text{p} = 
0.175$). Because lumbar puncture in VMAP is an optional procedure, 
the observed association between missingness and sex suggests that 
the missingness mechanism may be consistent with MAR rather than 
MCAR. Nevertheless, MAR cannot be verified using observed data alone, 
and the possibility of MNAR due to unobserved clinical or behavioral 
factors cannot be ruled out.

\begin{table}[tbp]
	\centering
	\small
	\renewcommand{\arraystretch}{1.25}
	\setlength{\tabcolsep}{18pt}
	\begin{tabular}{ll cccc}
		\toprule
		Parameter & Metric & \multicolumn{4}{c}{Missing Data Methods} 
		\\
		\cmidrule(lr){3-6}
		& & CCA & MICE & MLP & VAE \\
		\midrule 
		\multirow{3}{*}{Intercept} & EST & 32.13 & 32.47 & 33.86 & 
		33.74 \\
		 & SE & 3.02 & 2.21 & 1.93 & 1.92 \\
		 & P-Val & $< 0.01$ & $< 0.01$ & $< 0.01$ & $< 0.01$ \\
		 \midrule
		 \multirow{3}{*}{CSFRatio} & EST & 29.55 & 31.97 & 29.17 & 
		 30.88 \\
		 & SE & 9.82 & 10.87 & 10.25 & 10.06 \\
		 & P-Val & $< 0.01$ & $< 0.01$ & $< 0.01$ & $< 0.01$ \\
		 \midrule
		 \multirow{3}{*}{Age} & EST & $-0.12$ & $-0.13$ & $-0.15$ & 
		 $-0.15$ \\
		 & SE & 0.04 & 0.03 & 0.02 & 0.02 \\
		 & P-Val & $< 0.01$ & $< 0.01$ & $< 0.01$ & $< 0.01$ \\
		 \midrule
		 \multirow{3}{*}{Sex} & EST & $-0.09$ & 0.29 & 0.09 & 0.08 \\
		 & SE & 0.51 & 0.37 & 0.35 & 0.35 \\
		 & P-Val & 0.86 & 0.43 & 0.78 & 0.82 \\
		\bottomrule
	\end{tabular}
	\caption{Regression results from applying different missing data 
	imputation methods to the VMAP dataset.}
	\label{tab:vmap}
\end{table}

We regressed MoCA on CSF ratio, adjusting for age and sex. 
Table~\ref{tab:vmap} summarizes the results, suggesting that all four 
methods produce consistent findings: a higher CSF 
A$\beta_{42}$/A$\beta_{40}$ ratio is significantly associated with 
better cognitive performance, in line with existing literature. The 
consistency of these findings across different missing data 
approaches suggests that the estimated CSF--cognition association is 
relatively insensitive to the choice of method among those 
considered. The estimated intercepts and age effects are similar 
across methods with comparable standard errors. Although the sex 
effect is not statistically significant for any method, CCA suggests 
slightly higher MoCA scores for males, whereas MICE, MLP, and VAE 
estimate a small negative association. Among them, MICE produces the 
largest magnitude $0.29$, while MLP and VAE yield nearly identical 
estimates.

\section{Discussion}
\label{sec:dis}

In this study, we conduct a comprehensive evaluation of missing data 
imputation strategies including traditional statistical methods, 
machine learning approaches, and deep learning algorithms. Using 
extensive simulations under MCAR, MAR, and MNAR mechanisms with 
varying missing proportions, we find that VAE generally provides the 
most favorable balance between estimation bias and efficiency across 
the simulated scenarios, including the challenging MNAR setting where 
missingness depends on both the unobserved covariate and the observed 
outcome. MICE remains competitive under MCAR and MAR, offering 
improved efficiency relative 
to CCA. Other deep learning approaches, such as MLP, also perform 
reliably when MNAR does not depend on the outcome. In contrast, 
GAN-based imputation exhibits greater estimation variability and 
discrepancies between empirical standard deviations and estimated 
standard errors, raising concerns about inferential reliability. 
Other machine 
learning methods (e.g., kNN, SVM and RF) show greater susceptibility 
to bias, especially for estimating key biomarker effects. Overall, 
our findings highlight the importance of aligning imputation strategy 
with both the underlying missingness mechanism and the scientific 
objective.

From a practical perspective, the choice of imputation method should 
consider the underlying missingness mechanism, sample size, 
computational requirements, and inferential objectives. For 
biomedical studies where MAR is considered plausible, MICE remains a 
reliable and accessible approach, particularly when uncertainty 
quantification is important. MLP and VAE provide promising 
alternatives, with VAE demonstrating competitive performance across 
our simulation settings. In contrast, the relatively poor performance 
of kNN, SVM, and RF in our study suggests that their application to 
regression inference requires careful consideration of model 
specification and hyperparameter tuning. When MNAR is suspected, 
researchers should exercise additional caution, as no imputation 
method can universally guarantee unbiased inference without 
additional assumptions.

Despite the strengths of our study, several limitations point to 
future research opportunities. First, our simulations focus on a 
cross-sectional linear model with moderate sample size; performance 
may differ in longitudinal or high-dimensional settings where 
nonlinear relationships and temporal dependence are present. 
Second, the relatively poor performance of kNN and RF may reflect 
limited hyperparameter tuning in standard \proglang{R} 
implementations, and more rigorous model tuning may lead to
improvements. Third, although DL-based methods demonstrated promising 
performance, their effectiveness may depend on sample size, network 
architecture, and computational resources. Future research should 
further investigate their scalability and robustness across more 
complex biomedical data structures. Fourth, uncertainty 
quantification for deep learning imputation methods remains 
underdeveloped, motivating our future investigation into principled 
inference and variability assessment for these models. Finally, the 
missingness mechanism in the VMAP application cannot be definitively 
established from observed data alone, and the robustness of our 
findings to potential departures from MAR warrants further 
investigation through sensitivity analyses.

\section*{Acknowledgments}

VMAP is supported by NIH fund: R01-AG034962. Dr.\ Xie is supported 
by NIH fund: P30-AG072979, P01-AG066597, P01-AG084497.

\bibliographystyle{chicago}
\bibliography{refs}
	
\end{document}